\documentclass{article}

\usepackage{PRIMEarxiv}

\usepackage[utf8]{inputenc} 
\usepackage[T1]{fontenc}    
\usepackage[hidelinks]{hyperref}       

\hypersetup{
    colorlinks=true,
    linkcolor=blue,
    citecolor=blue,
    urlcolor=blue
}

\usepackage{url}            
\usepackage{booktabs}       
\usepackage{amsfonts}       
\usepackage{nicefrac}       
\usepackage{microtype}      
\usepackage{lipsum}
\usepackage{fancyhdr}       
\usepackage{graphicx}       
\graphicspath{{media/}}     
\usepackage{type1cm}        
\usepackage{makeidx}         
\usepackage{graphicx}        
\usepackage{multicol}        
\usepackage[bottom]{footmisc}

\usepackage{newtxtext}       %
\usepackage[varvw]{newtxmath}       
\usepackage{csquotes}
\usepackage{tcolorbox}

\def\eg{, e.g.}
\def\ie{, i.e.}
\title{Machine Learning in Biomechanics: Key Applications and Limitations in Walking, Running, and Sports Movements
\thanks{\textbf{This is a preprint of the following chapter:} Carlo Dindorf, Fabian Horst, Djordje Slijepčević, Bernhard Dumphart, Jonas Dully, Matthias Zeppelzauer, Brian Horsak, and Michael Fröhlich, \textit{Machine Learning in Biomechanics: Key Applications and Limitations in Walking, Running and Sports Movements}, published in \textit{Artificial Intelligence, Optimization, and Data Sciences in Sports}, edited by Blondin, M.J., Fister Jr., I., and Pardalos, P.M., 2025, Springer, Cham, reproduced with permission of Springer. The final authenticated version is available online at: \href{https://doi.org/10.1007/978-3-031-76047-1_4        }{https://doi.org/10.1007/978-3-031-76047-1          \_4}.}
}

\author{ 
  Carlo Dindorf$^\Delta$ \\
  Department of Sports Science \\
  University of Kaiserslautern-Landau (RPTU) \\
  Kaiserslautern \\
  \texttt{carlo.dindorf@rptu.de} \\
  \And
  Fabian Horst$^\Delta$ \\
  Institute of Sports Science \\
  Johannes Gutenberg-University Mainz \\
  Mainz \\
  \texttt{horst@uni-mainz.de} \\
  \And
  Djordje Slijepčević$^\Delta$ \\
  Institute of Creative Media Technologies \\
  St. Pölten University of Applied Sciences \\
  St. Pölten \\
  \texttt{djordje.slijepcevic@fhstp.ac.at} \\
  \And
  Bernhard Dumphart \\
  Institute of Health Sciences \& Center for Digital Health and Social Innovation \\
  St. Pölten University of Applied Sciences \\
  St. Pölten \\
  \texttt{bernhard.dumphart@fhstp.ac.at} \\
  \And
  Jonas Dully \\
  Department of Sports Science \\
  University of Kaiserslautern-Landau (RPTU) \\
  Kaiserslautern \\
  \texttt{jonas.dully@rptu.de} \\
  \And
  Matthias Zeppelzauer$^\Psi$ \\
  Institute of Creative Media Technologies \\
  St. Pölten University of Applied Sciences \\
  St. Pölten \\
  \texttt{matthias.zeppelzauer@fhstp.ac.at} \\
  \And
  Brian Horsak$^\Psi$ \\
  Institute of Health Sciences \& Center for Digital Health and Social Innovation \\
  St. Pölten University of Applied Sciences \\
  St. Pölten \\
  \texttt{brian.horsak@fhstp.ac.at} \\
  \And
  Michael Fröhlich$^\Psi$ \\
  Department of Sports Science \\
  University of Kaiserslautern-Landau (RPTU) \\
  Kaiserslautern \\
  \texttt{michael.froehlich@rptu.de} \\
  \\
  $^\Delta$ Shared first authorship; authors contributed equally; $^\Psi$ Shared senior authorship; authors contributed equally}

\begin{document}
\maketitle

\begin{abstract}
This chapter provides an overview of recent and promising Machine Learning applications\ie~\textit{pose estimation}, \textit{feature estimation}, \textit{event detection}, \textit{data exploration \& clustering}, and \textit{automated classification}, in gait (walking and running) and sports biomechanics. It explores the potential of Machine Learning methods to address challenges in biomechanical workflows, highlights central limitations\ie~\textit{data and annotation availability} and \textit{explainability}, that need to be addressed, and emphasises the importance of interdisciplinary approaches for fully harnessing the potential of Machine Learning in gait and sports biomechanics.
\end{abstract}

\vspace{0.8cm}


\section{Introduction}
\label{sec:intro}

Artificial Intelligence~(AI), specifically Machine Learning~(ML), has attracted increasing interest in many domains, including biomechanics. The biomechanics domain focuses on the quantitative description and analysis of human movements, taking into account aspects such as kinematics, kinetics, and electromyography~\cite{Winterbiomechanics.2009}. This comprehensive approach integrates diverse data sources to gain insights into the intricate mechanics of human movement, including gait\footnote{The term ``gait'' is utilised when referring to both walking and running movements throughout this chapter.} and sports biomechanics. The motivation for using ML in biomechanics is driven, among other things, by two central challenges in biomechanical workflows: 

\textbf{Laboratory-Bound, Resource and Time-Intensive Data Acquisition and Processing.} In the biomechanics domain, there is often a need for a precise and comprehensive understanding of human movement, encompassing various body parts and involving a range of biomechanical signals such as joint angles, ground reaction forces, and muscular activities. Ensuring high temporal and spatial precision of the data typically requires standardised laboratory-bound setups with expensive measurement equipment\eg~marker-based infrared camera systems and force platforms as well as skilled personnel. As a consequence, acquiring and processing biomechanical data can be resource-intensive and time-consuming. First, ML can be employed to assist domain experts in data acquisition, for example by estimating biomechanical signals through simple sensor recordings, such as inertial measurement units~(IMUs) or video images, even in natural environments\eg~outside the laboratory. Second, ML can aid in data processing, for example in data segmentation, as well as identification of measurement errors and outliers. The overarching goal in employing ML is to accelerate biomechanical workflows and make biomechanical data more accessible while simultaneously enhancing data quality. 

\textbf{Complexity of Data Analysis.} Biomechanical data are extensive and encompass multi-dimensional and multi-correlated time series signals from various body parts~\cite{chau.2001}. Biomechanical signals are characterised by a substantial level of both inter-subject~\cite{horst2023modeling} and intra-subject~\cite{horst.2016} variability, alongside the presence of numerous interacting influential factors. For example, when assessing a pathology through ground reaction forces during walking, it is essential to take into account that these forces can be influenced not only by the presence of an underlying pathology, but also by the age \cite{slijepcevic2022explaining}, sex \cite{horst2020explaining}, body mass, footwear, and walking speed \cite{horst2024explainable} of an individual. Consequently, during data analysis even experienced domain experts face challenges in interpreting the extensive and complex biomechanical data at hand. The integration of ML methods for data exploration, clustering, and automated classification seems promising in supporting and enhancing biomechanists' efforts in data analysis and interpretation. Ultimately, leveraging ML holds the potential to provide data-driven decision support in both research and applied settings.

ML approaches have proven to be effective in handling complex data and automating various aspects of biomechanical data acquisition, processing, and analysis. These approaches hold the potential to enhance the efficiency and accuracy of biomechanical workflows and provide biomechanists with valuable tools for gaining deeper insights into human movement. With the integration of ML, the biomechanics domain stands on the brink of substantial progress in advancing our understanding of human movement and its wide-ranging applications.

The present chapter aims to give an overview of promising applications of ML in the biomechanics domain and elaborates central challenges and limitations. In Section~\ref{sec:background}, we provide the reader with background information describing usual biomechanical (Section~\ref{sec:biomechanics}) and ML (Section~\ref{sec:machinelearning}) workflows. In Section~\ref{sec:applications}, each subsection offers a description and related work of the selected ML applications in biomechanics, while Section~\ref{sec:limitations} delves into the overarching central challenges and limitations. Throughout this chapter, we will provide separate descriptions for applications in the biomechanics of walking, running, and sports movements. To conclude this chapter (Section~\ref{sec:conclusion}), we consolidate valuable insights from these movements and identify potential synergies for future progress in the domain.

\section{Biomechanics and Machine Learning Workflows}
\label{sec:background}
\subsection{Biomechanics Workflow}
\label{sec:biomechanics}

Biomechanical testing typically follows a three-phase workflow: data acquisition, data processing, and data analysis. In the following, we will outline each of these phases and motivate how ML applications can assist and enhance the capabilities of biomechanists throughout this workflow. Comprehensive background information can be found in biomechanics textbooks~\cite{Winterbiomechanics.2009, uchida2021biomechanics}.

\subsubsection{Data Acquisition}
\label{sec:biomechanics-acquisition}

\begin{figure*}
\centering
\includegraphics[width=1\linewidth]{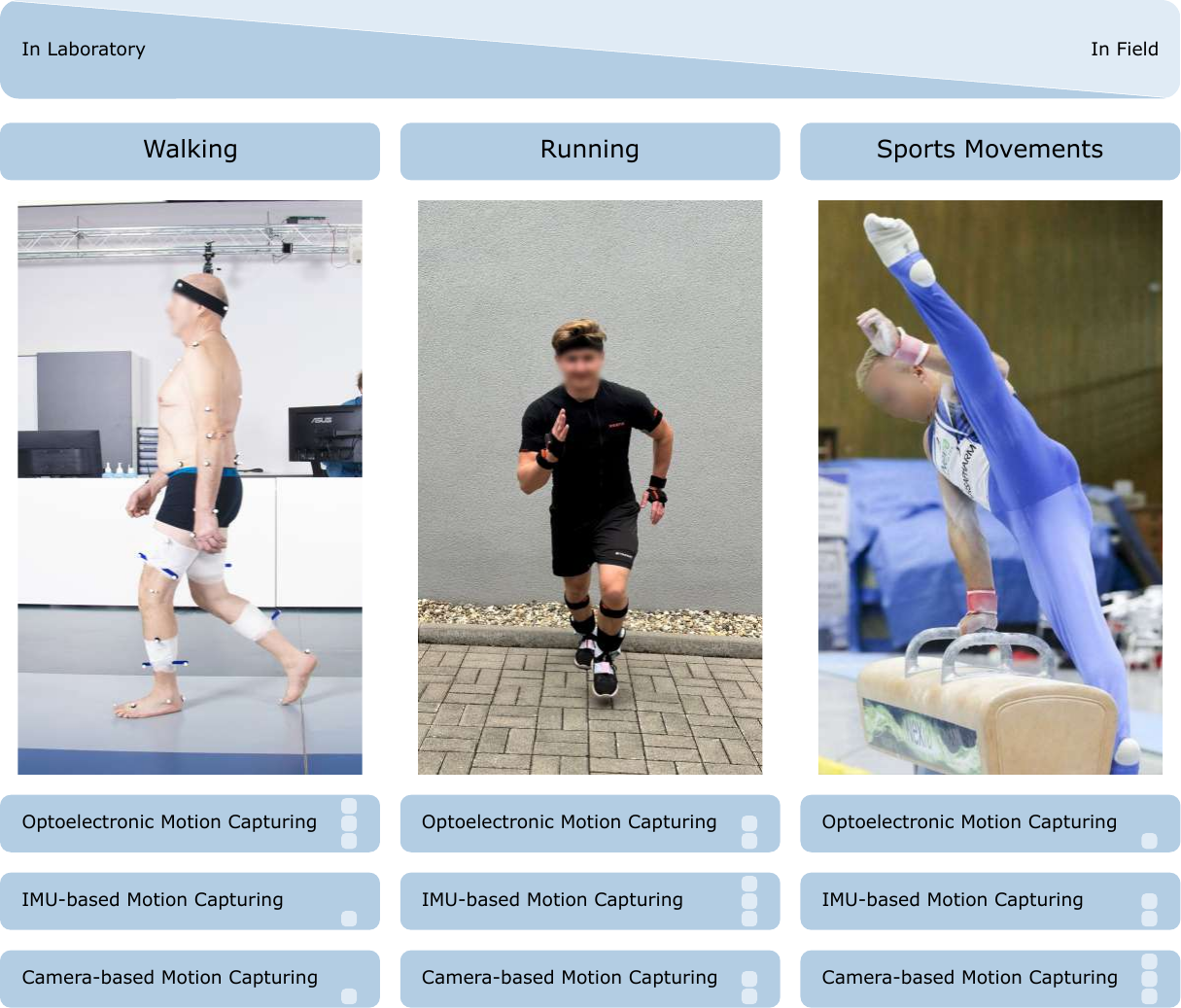}
\caption{Overview of three commonly applied biomechanical testing approaches: i) biomechanical testing using marker-based infrared camera systems together with force platforms and electromyography in laboratories can be considered the gold standard, where achieving the highest level of precision and accuracy are prioritised over material and personal resources; ii) biomechanical testing using wearable sensors\eg~IMUs in natural field-based settings, offering both long-term recordings and real-time feedback; and iii) biomechanical testing utilising portable sensors\eg~RGB video cameras is well-suited for conducting assessments in natural field-based settings, where the emphasis is on minimising constraints, even at the expense of spatial and temporal accuracy. The number of squares next to a measurement approaches describe its prevalence for the biomechanical testing of a movement task (walking, running, sports movements). Copyright (left photo): Helene Sorger / St.Pölten University of Applied Sciences.
}
\label{fig:testingapproaches}
\end{figure*}

Biomechanical data acquisitions involve the use of a variety of measurement devices to measure i) kinematic\eg~positions and accelerations of body parts, ii) kinetic\eg~ground reaction forces, iii) electromyographic\eg~muscle activity, and iv) anthropometric\eg~body mass, signals during human movements. 
In the following, we will outline common biomechanical testing approaches, including laboratory-based\eg~using marker-based infrared camera systems, force platforms, and electromyography, as well as field-based\eg~using wearable sensors such as IMUs and pressure insoles and portable sensors such as RGB video cameras approaches as exemplified in Figure~\ref{fig:testingapproaches}. 

\textbf{Laboratory-Based Approach.} In the realm of fundamental and clinical research exploring biomechanical principles in walking, running, and sports movements, biomechanists frequently choose to utilise standardised laboratory settings. Biomechanical testing in laboratory settings often involves analysing kinematic signals by tracking markers placed at specific anatomical landmarks on the body with multiple infrared cameras. This approach provides precise 3D position data of the body segments and enables\eg~the estimation of joint angles using physics-based biomechanical modelling. 
Force platforms and electromyography are frequently used alongside marker-based approaches. Force platforms provide insight into the magnitude, direction, and timing of forces exerted between the body and the platform. These platforms provide high temporal and spatial resolution, but require a proper physical contact with the platform. In certain applications, the electrical activity of muscles during movement is recorded to gain insights into muscle activation patterns using electromyography with electrodes attached to the skin. Comprehensive laboratory-based approaches allow biomechanical modelling that integrates the recorded marker positions, ground reaction forces, and muscular activities to estimate biomechanical signals that cannot be directly measured\eg~joint angles and moments, joint contact forces, and muscle forces. The analysis of all these signals enables a comprehensive and accurate understanding of movement mechanics. 

The laboratory-based approach provides high accuracy and precision, but also requires high personnel and material resources. Moreover, it comes along with certain experimental limitations. Measurement devices like infrared cameras are often bound to indoor environments without sunlight and cover only a limited spatial area, in which the movement analysis can take place. In addition, the attachment of markers and electrodes to the body may not be feasible for all types of movements and may limit the subjects' range of motion. Thus, measurement methods used in laboratory-based approaches cannot be applied for everyday situations in the field\eg~standing and walking in public transport, tele-rehabilitation, and sports competitions. Nevertheless, the laboratory-based approach can be considered the established standard for biomechanical testing in terms of accuracy, that is used as the best reference when exploring the accuracy of field-based approaches using wearable sensors and video cameras. 

\textbf{Field-Based Approach.} Field-based methods are used to study biomechanical principles associated with walking, running, and sports movements in natural environments. In field settings, biomechanical testing frequently incorporates the use of wearable sensors for prolonged data acquisitions and real-time feedback. Additionally, RGB video cameras are employed, eliminating the need for sensor attachment to the body.

Wearable sensors\eg~IMUs and pressure insoles are wireless and lightweight. While these sensors often provide less detailed and more noisy data than laboratory-based approaches, they can be used in various environments with less restricted experimental conditions. 
IMUs consist of a combination of accelerometers, gyroscopes, and sometimes magnetometers. IMUs can be attached to a body segment or equipment\eg~shoes to measure changes in acceleration and orientation. These data can be used to estimate kinematic signals\eg~joint angles in real-time in non-laboratory settings. This allows, for example, the analysis of running biomechanics in natural environments\eg~long distance running on diverse natural terrains such as forest trails. This analysis, however, is limited to kinematic signals like joint angles. Estimating kinetic signals commonly analysed in laboratory-based approaches, such as ground reaction forces and joint moments, presents challenges when relying only on IMU data. ML can offer a promising solution to this challenge, namely ML-based biomechanical \textbf{feature estimation}. For example, an ML method can be used to predict ground reaction forces from available IMU data and therefore significantly enhance the capabilities of IMUs in biomechanical research by providing accurate and reliable estimations of biomechanical signals.

\begin{tcolorbox}[width=\textwidth,
                  colback=gray!10,
                  colframe=black,
                  boxrule=1pt, 
                  arc=5pt,
                  boxsep=5pt,
                  left=10pt,
                  right=5pt,
                  top=5pt]
    \textbf{Key Application: Feature Estimation (Section~\ref{sec:feature})}
    
    Feature estimation employs ML to predict complex biomechanical data\eg~joint angles and moments from more accessible data sources, including IMUs, pressure insoles, and RGB video cameras.
\end{tcolorbox}

Despite the possibilities that wearable-based approaches offer for assessing movement in the field, several sports-specific movements pose difficulties in terms of attaching hardware to the body. This can include limitations on range of motion, which is especially relevant during competition, and the potential risk of injury to athletes who might fall on the utilised sensors\eg~high jump.

In situations where sensor and marker attachment to subjects' bodies is not feasible, such as during sports competitions, researchers often resort to acquiring image and video data using cameras, including video or depth cameras, for kinematic analyses in sports practice. In this process, biomechanists manually identify and digitise anatomical landmarks in each frame of a video. However, the accuracy of this manual digitisation depends heavily on factors like camera resolution and lighting conditions. Moreover, this procedure is very time-consuming, so that the analysis of a larger number of recordings of a competition or a training session is only practicable to a limited extent. ML offers a promising solution to address these challenges. Specifically, in the context of human \textbf{pose estimation}, ML algorithms can be trained to automatically recognise and track anatomical landmarks in video or image data. 

\begin{tcolorbox}[width=\textwidth,
                  colback=gray!10,
                  colframe=black,
                  boxrule=1pt, 
                  arc=5pt,
                  boxsep=5pt,
                  left=10pt,
                  right=5pt,
                  top=5pt]
    \textbf{Key Application: Pose Estimation (Section~\ref{sec:pose})}
    
    Pose estimation is the process of automatically tracking and determining the body's anatomical landmarks, body segments, or joint centre locations in video images using ML, enabling the quantification of human movement without marker and sensor attachments to the human body.
\end{tcolorbox}

Integrating ML into the biomechanical workflow allows for more extensive kinematic analyses, even in scenarios where hardware attachment is not feasible. Thus, the use of ML could significantly reduce the time and effort required to obtain and analyse biomechanical data. This would enable researchers, coaches, and clinicians to analyse a more comprehensive set of recordings from natural environments such as sports competitions or training sessions.

\subsubsection{Data Processing}
\label{sec:biomechanics-processing}
Data processing is critical for reducing sensor noise in biomechanical signals and making them comparable across different subjects and recordings. In the following, we briefly mention common processing steps: First, \textbf{filtering} is used to smooth out irregularities in the raw signals to enhance the clarity of the movement patterns for subsequent analysis. After filtering, the biomechanical signals undergo an \textbf{event detection} to identify and extract movement (sub-)phases in the recordings. 

\begin{tcolorbox}[width=\textwidth,
                  colback=gray!10,
                  colframe=black,
                  boxrule=1pt, 
                  arc=5pt,
                  boxsep=5pt,
                  left=10pt,
                  right=5pt,
                  top=5pt]
  \textbf{Key Application: Event Detection (Section~\ref{sec:event})}
  
  Event detection in time series data is the annotation of certain events that are used to extract useful and vital information or to remove unwanted and unnecessary data for further analysis.
\end{tcolorbox}

Afterwards, the integrity of the biomechanical signals is checked using \textbf{outlier detection} techniques. Identifying and removing outliers or artefacts due to sensor shifts or errors is essential to prevent them from distorting the data analysis. \textbf{Normalisation} involves scaling the signals based on specific criteria, often individual characteristics like body size, body mass, and speed. This adjustment allows for fair comparisons between different subjects and conditions, as it accounts for variations in size that can influence biomechanical data. Normalisation ensures that data is presented on a consistent scale, making it easier to draw meaningful comparisons and conclusions.
\subsubsection{Data analysis}
\label{sec:biomechanics-analysis}

The analysis of biomechanical data poses challenges due to the extensive volume of data recorded during measurement sessions, that are multidimensional and interrelated~\cite{chau.2001}. The following two analysis approaches are commonly employed for data analysis in practice:

\textbf{Qualitative Assessment.} This approach involves the visual inspection and qualitative assessment of biomechanical signals. Researchers examine the data to identify patterns, anomalies, or noteworthy events. While qualitative assessment can provide valuable insights into the data, it often lacks sufficient inter-rater reliability, leading to potential discrepancies in interpretation among different analysts~\cite{Eastlackinterrater.1991,Brunnekreefreliability.2005}.

\textbf{Extraction of Biomechanical Features.} Alternatively, biomechanists often choose to extract specific ``handcrafted'' features from the time series data. For instance, they might extract discrete features such as the maximal and minimal knee angle during walking. Subsequently, inferential statistical analysis is applied to these extracted features. However, the approach of extracting single time-discrete features from several multidimensional time series signals carries certain risks. Reducing complex time series data to a small set of discrete features can potentially lead to information loss, limiting the holistic understanding of biomechanical processes. Additionally, the selection of which features to extract and analyse can introduce subjectivity and bias into the analysis.

To address these challenges and enhance the reliability of biomechanical data analysis, researchers are exploring the integration of ML methods. ML models can be trained using any input data, including handcrafted features, automatically extracted features, or the raw time series signals. The latter approach eliminates the need for prior reduction of the available biomechanical time series data. This holistic approach enables the incorporation of comprehensive information, encompassing time series data of whole-body 3D joint angles and ground reaction forces throughout the entire movement cycle.

ML methods, such as Deep Learning models, can automatically identify relevant patterns and features within the multidimensional and interrelated data. This reduces the reliance on manual feature extraction and has the potential to improve the overall accuracy and objectivity of biomechanical analyses. Data exploration and clustering involves the use of unsupervised ML methods to uncover hidden patterns or clusters within the data. This aids in understanding the inherent structure of biomechanical data and can guide subsequent analyses.

\begin{tcolorbox}[width=\textwidth,
                  colback=gray!10,
                  colframe=black,
                  boxrule=1pt, 
                  arc=5pt,
                  boxsep=5pt,
                  left=10pt,
                  right=5pt,
                  top=5pt]
  \textbf{Key Application: Data Exploration and Clustering (Section~\ref{sec:cluster})}
  
    Clustering involves grouping instances or individuals with similar biomechanical characteristics using unsupervised ML, thereby revealing underlying structures and subgroups within complex biomechanical data.
\end{tcolorbox}

In addition to unsupervised ML, supervised ML plays a crucial role in biomechanical data analysis. Supervised ML leverages annotated data to address a wide range of classification and regression tasks in biomechanical settings. Supervised ML can be employed to automatically identify and categorise\eg~specific pathological conditions.

\begin{tcolorbox}[width=\textwidth,
                  colback=gray!10,
                  colframe=black,
                  boxrule=1pt, 
                  arc=5pt,
                  boxsep=5pt,
                  left=10pt,
                  right=5pt,
                  top=5pt]
  \textbf{Key Application: Automated Classification (Section~\ref{sec:classification})}
  
    Automated classification refers to the process of developing a predictive model that assign input features of data samples to predefined categories or classes using supervised ML.
\end{tcolorbox}

The use of ML, encompassing data exploration, clustering through unsupervised ML methods, and automated classification and regression via supervised ML methods, holds promise for advancing our understanding of human movement.

\subsection{Machine Learning Workflow} 
\label{sec:machinelearning}

\begin{figure*}
\centering
\includegraphics[width=1\linewidth]{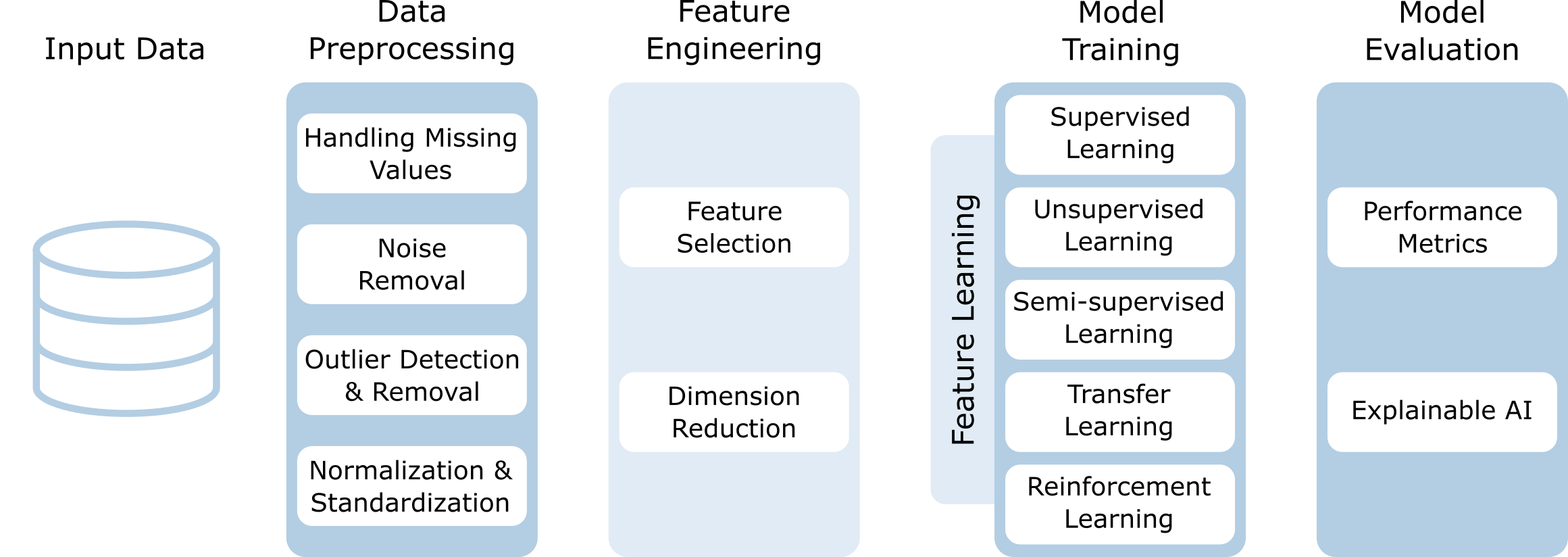}
\caption{In the realm of biomechanics, typical ML workflows are characterised by the following phases: data preprocessing, optionally feature engineering, ML model training (which can incorporate feature learning), and ML model evaluation.}
\label{fig:MLWorkflow}
\end{figure*}

ML approaches have the ability to model relationships between versatile input and output data, enabling various applications throughout the biomechanical workflow. In the following, we will outline the typical ML workflow and emphasise the requirements for applying ML in biomechanics. For more in-depth explanations, refer to relevant sources like~\cite{Richter.2021}. The typical ML workflow can be characterised by the following phases~(Figure~\ref{fig:MLWorkflow}): data preprocessing, feature engineering, ML model training, and ML model evaluation.

\subsubsection{Machine Learning-Specific Data Preprocessing}

Data preprocessing is an essential step in preparing data for ML models. When dealing with biomechanical data, \textbf{data cleaning} is usually omitted, because in the domain of biomechanics, data is already typically filtered, normalised to account for various influencing factors including walking speed and body mass, and outliers are removed, prior to data analysis~(Section~\ref{sec:biomechanics-processing}). Hence, these data cleaning steps are only necessary when applying ML to raw sensor signals.
 
\textbf{Data Normalisation and Standardisation} become necessary when signals/features in a dataset reside within different value ranges\eg~different value ranges of joint angles and joint moments. A major motivation for this preprocessing step is to ensure that each signal/feature has an equal opportunity to contribute to the ML model's predictions. Furthermore, in the context of Deep Learning, data normalisation is usually leveraged to foster the convergence of gradient descent during training. Common techniques include min-max normalisation, z-score standardisation, and transformations like logarithmic or polynomial transformations. The choice of method can vary based on the respective data and domain. Recent studies showed that these techniques significantly impact model performance~\cite{burdack.2020} and behaviour~\cite{slijepcevic2021explaining}, however, there are very few studies that systematically compare these techniques in the context of biomechanical data.

\subsubsection{Feature Engineering Versus Feature Learning}
Feature engineering involves extracting new features from the preprocessed data to capture more meaningful information for the specific task at hand. Feature engineering is crucial not only for improving model accuracy, preventing overfitting, and reducing computational demands but also for enhancing interpretability and extracting meaningful information~\cite{Liu.2005}. 
In conventional methods, feature engineering typically encompasses the crafting of domain-specific features by domain experts. These features are often derived as time-discrete features, such as local and global minima, maxima, or other simple descriptive statistics. Moreover, feature selection methods can be employed using ML to automatically select a subset of features from the initially defined feature set. Another commonly applied approach in biomechanics leverages dimension reduction techniques to extract features. Dimension reduction aims to project input data into a lower-dimensional feature space, thereby effectively removing redundant and irrelevant information.

However, it has also been demonstrated that these simple feature engineering approaches can lead to loss of information and performance of the ML models~\cite{slijepcevic.2017}. In Deep Learning, we refer to feature learning, indicating that the process of feature engineering is replaced, and instead, the Deep Learning model takes over the task of learning robust features from the data. As Deep Learning methods can automatically learn task-relevant features from the data during the training process, they hold the potential to outperform the aforementioned traditional approaches~\cite{Dindorf.2021extraction}. 

\subsubsection{Model Training}

Depending on the input data and the objectives of the task, different strategies can be selected to train ML models. Some of the main learning strategies already utilised in biomechanics, or those that may become more significant, encompass:

\textbf{Supervised Learning.} An ML algorithm is trained on an annotated\ie~labelled dataset, where each data sample contains input features and corresponding target labels. The goal is to model relationships between the input features and target labels so that the ML model can make accurate predictions on unseen data for classification (Section~\ref{sec:classification}) or regression\eg~feature estimation (Section~\ref{sec:feature}) tasks.

\textbf{Unsupervised Learning.} An ML method utilises non-annotated\ie~unlabelled data to discover patterns, clusters, or relationships within the data based on structural properties of the data. Common applications include clustering similar data points or reducing the dimensionality of available data (Section~\ref{sec:cluster})\eg~for data exploration.

\textbf{Semi-Supervised Learning.} This training approach combines elements of both supervised and unsupervised learning. This approach leverages first a larger amount of unlabelled data to model structural characteristics of the data in an unsupervised way and then utilises a small amount of labelled data to model the task-specific target function.

\textbf{Transfer Learning.} This training approach involves pre-training a model on a large dataset and fine-tuning it for a specific task on a smaller dataset. The goal is to leverage patterns learned from one task or domain to improve performance on a related task with limited data.

\textbf{Reinforcement Learning.} An agent learns to make a sequence of decisions or actions within an environment through a process of trial and error, aiming to maximise a cumulative reward received as feedback from that environment.

\subsubsection{Model Evaluation}
Model evaluation is a crucial step in ML, involving the assessment of model performance through performance metrics and enhancing model transparency through explainability methods. These methods enable ML models to be interpretable to users, which is a crucial requirement for many biomechanical applications and regulatory compliance, in addition to their overall performance.

\textbf{Evaluating Model Performance} primarily involves assessing the generalisation ability of a trained ML model on unseen data. There are several evaluation strategies such as hold-out, k-fold cross-validation, and leave-one-out cross-validation. The selection of the approach is commonly influenced by the quantity of available data and the specific domain of application. Regardless of the approach, it is important to maintain the integrity of the test data\eg~for the task of classifying gait pattern to ensure that each subject's data is exclusively assigned to either the training data or test data. Performance metrics are quantitative measures used to assess how well an ML model performs on a specific task. These metrics vary depending on the type of problem\eg~classification, regression, clustering and the specific objectives of the task.

\textbf{Explainable Artificial Intelligence (XAI)} involves methods that aim to enhance the transparency and interpretability of ML models. Considering the inherent ``black-box'' nature of complex ML and Deep Learning models, explainability methods allow users, stakeholders, and regulators to gain insight into the model behaviour and decisions. For a comprehensive overview of XAI in biomechanics, refer to Section~\ref{subsec:explainability}.

\section{Key Applications of Machine Learning in Biomechanics}
\label{sec:applications}

\begin{figure*}
\centering
\includegraphics[width=1\linewidth]{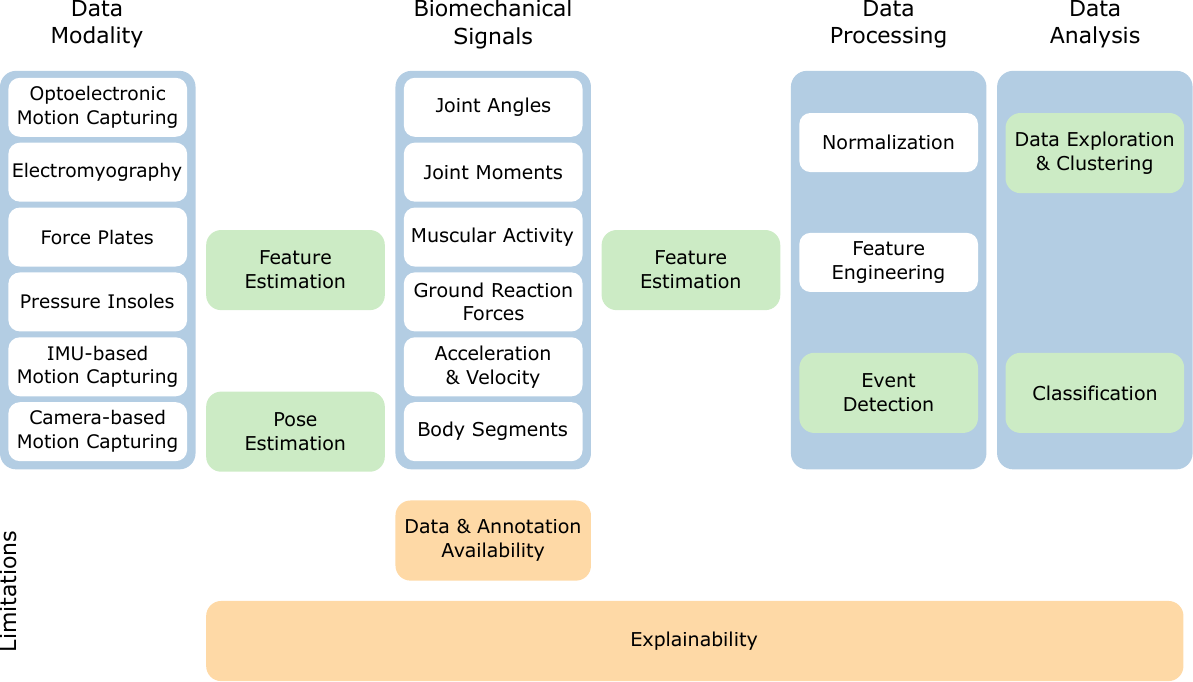}
\caption{Overview of the key ML applications (green boxes)\ie~pose estimation, feature estimation, event detection, data exploration \& clustering, and automated classification, and their integration into the phases of the biomechanical workflow. The key limitations (orange boxes) associated with the use of ML in biomechanics include data and annotation availability and explainability.}
\label{fig:applications}
\end{figure*}

\begin{figure*}
\centering
\includegraphics[width=1\linewidth]{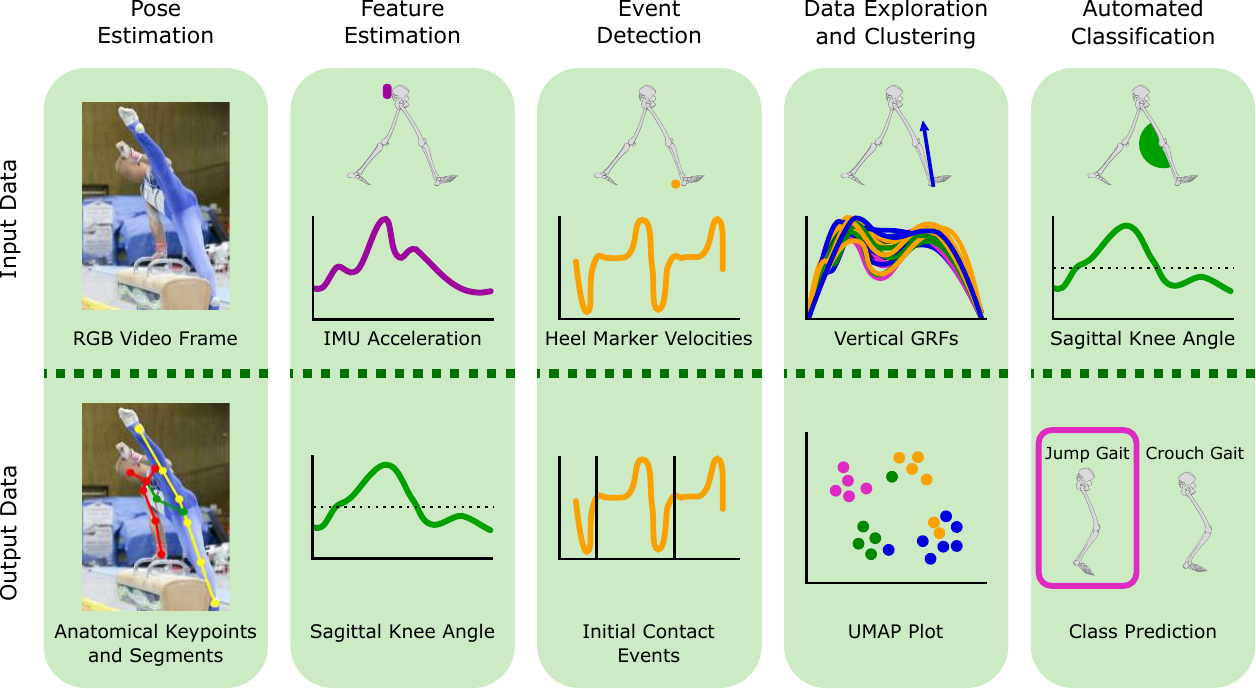}
\caption{Examples of input and output pairs for the key ML applications\ie~pose estimation, feature estimation, event detection, data exploration \& clustering, as well as automated classification.}
\label{fig:applications_detailed}
\end{figure*}

In the following, we showcase promising ML applications for key tasks in the biomechanical workflow. These applications encompass \textbf{pose estimation} (Section~\ref{sec:pose}), \textbf{feature estimation} (Section~\ref{sec:feature}), \textbf{event detection} (Section~\ref{sec:event}), \textbf{data exploration and clustering} (Section~\ref{sec:cluster}), as well as \textbf{automated classification} (Section~\ref{sec:classification}). Figure~\ref{fig:applications} provides an overview of the localisation of the key ML applications within the biomechanical workflow, while Figure~\ref{fig:applications_detailed} illustrates examples of input and output pairs. For each application we present a brief description of the task, followed by prominent research works, and conclude with application-specific limitations and future perspectives. Throughout this chapter, we will provide separate descriptions for applications in biomechanics related to walking, running, and sports movements.

\subsection{Pose Estimation}
\label{sec:pose}
\subsubsection{Problem Description}
Over the last three decades, marker-based infrared camera systems have become paramount in accurately quantifying and analysing human movement in sports and clinical settings for both research and decision-making purposes. They allow for comprehensive investigation of sports movements, gait, and functional activities, providing valuable insights to coaches, clinicians, and researchers that significantly advanced our understanding of the biomechanics of human movement. However, despite their value in human movement science and biomechanical research, their accessibility and widespread use are severely limited by high costs, complex hardware, time-consuming laboratory setups, and the need for highly trained operators.
Recent advancements in Deep Learning and computer vision technology offer a potential solution to these problems\ie~markerless motion capture. This technique allows the two- and three-dimensional tracking of human locomotion from simple videos. Markerless motion capture systems typically utilise Deep Learning models trained on large-scale datasets containing hundred thousands of manually annotated images of individuals engaged in daily activities and sports movements. These so-called pose estimation algorithms can detect anatomical landmarks or joint centre locations automatically in videos that are used to derive the pose of the individual in each video frame or image.

\subsubsection{Approaches}
Various pose estimation algorithms have been published over the past decade such as OpenPose~\cite{8765346}, DeepLabCut~\cite{mathis_deeplabcut_2018}, DeepPose~\cite{DeepPose_2014}, DeeperCut~\cite{insafutdinov_deepercut_2016}. Popular pose estimation algorithms for biomechanical applications especially include OpenPose~\cite{cao2017realtime} and DeepLabCut~\cite{mathis_deeplabcut_2018}. OpenPose is known for its ability to track multiple individuals in images and its easy handling, while DeepLabCut allows users to retrain pre-trained algorithms for specific tasks, which can be especially useful for uncommon movements\eg~pathological gait or sports movements. For an in-depth review of pose estimation algorithm designs, readers are directed to the numerous reviews available~\cite{ben_gamra_review_2021,chen_monocular_2020,song_human_2021,dang_deep_2019,colyer_review_2018,cronin_using_2021}.

The field of markerless motion capture is rapidly advancing, with various approaches being recently introduced. These approaches can potentially be categorised in two broad main groups: monocular markerless motion capture which uses a single camera, and multi-camera markerless motion capture which uses video data from two or more synchronised cameras for pose estimation.

\textbf{2D Single-Camera Markerless Systems} utilise 2D pose estimation algorithms to extract joint centre locations from a single image or video. Accurate determination of relevant 2D planar joint angles relies on the assumption of perfect alignment of the camera with the plane of movement. Several studies have already evaluated concurrent validity of 2D markerless motion capture by comparing obtained results to a marker-based system~\cite{cronin_markerless_2019,moro_markerless_2020,drazan_moving_2021,serrancoli_marker-less_2020}. In these studies, activities analysed include walking, underwater running, jumping, and cycling. Joint centre locations differed less than 20 mm to the marker-based system and joint angles as well as spatiotemporal parameters were approximately similar. However, depending on the algorithm used, other studies reported greater differences in joint angles and joint centre locations~\cite{stenum_two-dimensional_2021,wade_applications_2022}. Most of the studies also reported that in some cases manual corrections were necessary for incorrectly identified joint centre locations due to\eg~body part occlusion and labelling of the wrong body side. In summary, 2D single-camera methods can provide reasonably accurate spatiotemporal parameters and planar 2D joint angles, but they are limited to the analysis of one body side only and an optimal camera positioning relative to the movement plane is assumed. There are also efforts in using single-camera markerless motion capture to obtain 3D pose estimation. However, errors for these methods are still high, with joint centre differences reported in the literature in the range of 30-60 mm, and thus, to date do not offer enough accuracy to aid biomechanical applications~\cite{SarafianosError.2016}. For a thorough review of technological aspects for 3D single-camera markerless systems the reader is referred to recent reviews~\cite{chen_monocular_2020,ji_survey_2020}.

\textbf{3D Multi-Camera Markerless Systems} utilise a minimum of two cameras. The multi-camera approach intends to reduce the problem of partly occluded body parts and typically utilises triangulation methods, such as the Direct Linear Transformation~\cite{hartley_triangulation_1997}, to estimate the 3D pose from the 2D anatomical key features identified in each single video or image. Compared to their single-camera counterparts, multi-camera systems generally seek to offer an alternative method to 3D marker-based state-of-the-art motion capture systems~\cite{wade_applications_2022}. A few studies have assessed the performance of multi-camera 3D markerless systems utilising\eg~the OpenPose pose estimation algorithm. These studies, conducted by Nakano et al.~\cite{nakano_evaluation_2020}, Slembrouck et al.~\cite{slembrouck_multiview_2020}, and Zago et al.~\cite{zago_3d_2020}, have reported variations in joint centre location ranging from 10 to 50 mm. It was observed that slower movements exhibited more accurate results, with average differences in joint centre location compared to marker-based methods ranging from 10 to 30 mm. Conversely, faster movements such as jumping and throwing yielded differences of 20 to 40 mm~\cite{nakano_evaluation_2020}, which were further exacerbated when captured at lower video frame rates~\cite{slembrouck_multiview_2020,zago_3d_2020}. Recently, Uhlrich et al.~\cite{uhlrich_opencap_2023} have proposed the open-source tool, OpenCap. It combines a multi-camera OpenPose approach with constrained musculoskeletal modelling in a web-based service to quantify human movement dynamics from a minimum of two calibrated smartphone videos. Uhlrich et al.~\cite{uhlrich_opencap_2022} and Horsak et al.~\cite{horsak_concurrent_2023} reported a difference in joint kinematics between OpenCap and a marker-based motion capturing for various activities including pathological walking patterns of approximately 5-6 degrees on average. Further a recent study by Horsak et al. \cite{horsak2024repeatability} found almost comparable test-retest reliability of OpenCap compared to marker-based systems, thereby further underscoring the potential of markerless systems in biomechanics.

While the aforementioned studies have used open-source algorithms such as OpenPose, in recent years also commercial systems have emerged, such as KinaTrax (KinaTrax, Inc., Boca Raton, FL), SIMI (Simi Reality Motion Systems, Unterschleißheim, Germany), Theia3D (Theia Markerless, Inc., Kingston, Ontario, Canada), and DARI Motion (DARI Motion, Overland Park, Kansas). However, given their very recent availability, there is currently a lack of peer-reviewed research for most of the systems. As of 2024, Theia3D and KinaTrax seem as the most popular system in biomechanical research with at least a few studies evaluating the reliability and concurrent validity for tasks like heel raising, stepping down, walking, hopping, squatting, and running-and-cutting manoeuvres~\cite{kanko_assessment_2021,kanko_inter-session_2021,kanko_concurrent_2021,wren_comparison_2023,ito_markerless_2022,song_markerless_2023, ripic_validity_2023}. While Kanko et al.~\cite{kanko_concurrent_2021} and Ripic et al.~\cite{ripic_validity_2023} reported accurate estimations of spatiotemporal parameters for walking using the Theia markerless system, Kanko et al.~\cite{kanko_assessment_2021,kanko_concurrent_2021} found differences of 22 to 36 mm for joint centre locations and 3 to 11 degrees for flexion/extension and abduction/adduction in walking kinematics, although rotations about the long axis of a body segment were higher ranging from 7 to 13 degrees. Song et al.~\cite{song_markerless_2023} investigated eight different functional movement tasks and found their results to be generally in agreement with the results of Kanko et al. \cite{kanko_concurrent_2021}, but observed increased differences with faster movements\eg~while running. Wren et al.~\cite{wren_comparison_2023} recently reported errors of approximately 5 to 6 degrees for the lower-limb joint kinematics in patients with a range of diagnose including cerebral palsy and club foot deformities. Further, Kanko et al.~\cite{kanko_inter-session_2021} also found the reliability of the Theia system being comparable or having even partly slightly superior reliability in walking kinematics compared to previously reported marker-based methods~\cite{manca_repeatability_2010,mcginley_reliability_2009}. Lastly, a recent study by Horsak et al.~\cite{horsakIntertrialVariabilityHigher2024} demonstrated that markerless pose estimation pipelines can introduce additionally variability in kinematic data and, therefore, recommended using averaged waveforms rather than single ones to mitigate this problem. Further, they highlighted that caution is advised when using variability-based metrics in gait and human movement analysis based on markerless data as increased inter-trial variability can potentially lead to misleading results. 

In summary, multi-camera markerless motion capturing appears to be a highly promising alternative to record 3D human movement without the need for attaching markers. However, while markerless multi-camera systems produce similar kinematic patterns, errors are still above desirable clinical thresholds of two to five degrees in several cases.

\textbf{Application Examples for Assessing Health and Performance.} The range of applications for pose estimation is constantly growing. From a human movement science perspective, examples cover the full range life span, from human development, performance optimisation, and injury prevention, to motor assessment of persons with motor impairments~\cite{stenum_applications_2021}. For example, Ossmy and Adolph~\cite{ossmy_real-time_2020} used pose estimation to quantitatively track the development of infant locomotion to monitor body movements. Their results highlighted potential for early detection and intervention for delays in locomotor development. Pose estimation also has been successfully used for the early detection and diagnosis of congenital movement-based disorders such as cerebral palsy. Investigations demonstrate promising results in the early detection of cerebral palsy based on automatic movement assessment from infant video recordings, with performance comparable to standardised cerebral palsy risk measures~\cite{adde_early_2010,rahmati_video-based_2014}. 

Pose estimation also has a wide range of applications that contribute to optimising human health and performance. Examples range from the identification of abnormal gait patterns during walking and running~\cite{chaaraoui_abnormal_2015,guo_3-d_2019,kondragunta_gait_2020}, fall detection~\cite{bian_fall_2015,chen_fall_2021}, assessment of movement quality in manual labour work settings~\cite{han_empirical_2013,han_vision-based_2013}, and assessment of sports-related injury risk, such as anterior cruciate ligament rupture~\cite{Blanchard_2019}. 

While the application of pose estimation in sports seems obvious for performance assessment, injury risk screening and biofeedback training, to date peer-reviewed studies with application examples are still scarce. Recent applications of pose estimation in sports include the estimation of boxing poses~\cite{lin_model_2023,lahkar_accuracy_2022}, the analysis of throwing kinematic during baseball pitching~\cite{fleisig_comparison_2022}, analysing ski jumping techniques~\cite{bao_pose_2023}, as well as the application for various sports-related movements such as squatting, running, or jumping~\cite{uhlrich_opencap_2022,song_markerless_2023, pagnon_pose2sim_2022}.

\subsubsection{Limitations and Future Perspectives}
Markerless motion capture holds exciting potential for advancing real-world biomechanical assessment for both, clinical settings and sports. It has not only the potential to make measurement of human movement substantially more accessible to the broad public, it allows to streamline movement analysis by reducing the time required for data acquisition and processing, without sacrificing reliability~\cite{kanko_inter-session_2021, horsak2024repeatability}. Additionally, markerless techniques offer greater flexibility in analysing the data and can even capture movement data while individuals are wearing clothes~\cite{horsak2024repeatability,keller_clothing_2022}, an aspect which is of particular interest for various applications in sports.

\textbf{Accuracy of Kinematic Measurements.} Research shows that in clinical settings, markerless methods generally yield similar results to marker-based motion capture in terms of spatiotemporal measures, but they still need improvements in accurately determining joint kinematics before they can support decision making.

\textbf{Noisy Data in Sports Applications.} While their application in sports seems obvious, the high movement speeds, cluttered background, various lightning conditions, and variations in shape due to worn equipment, present significant challenges for pose estimation in sports movements.

However, with the currently seen rapid developments in pose estimation algorithms, and the growing amount of annotated data being shared openly, improvements in the accuracy of markerless 3D motion analysis for both, clinical applications and sports, are likely to happen. In some scenarios one could argue that having estimated quantitative data is better than having no data\eg~in settings where exact kinematic and kinetic measurements are almost impossible to acquire, like in pole vaulting or ski jumping.

\subsection{Feature Estimation}
\label{sec:feature}

\subsubsection{Problem Description}

Advancements in biomechanical modelling have greatly improved our understanding of movement mechanics and dynamics over the past decades. However, achieving comprehensive biomechanical modelling with accurate estimation of joint angles and moments typically requires extensive data acquisitions in standardised laboratory environments, which are both time and resource-consuming. One goal in the biomechanics domain is the estimation of complex biomechanical signals with minimal experimental constraints and effort, even in natural environments outside of the laboratory. The use of ML approaches for feature estimation of complex biomechanical signals from simplified data acquisition and data processing pipelines has been increasingly explored in recent years.

\subsubsection{Approaches}

Feature estimation has predominantly been explored in gait biomechanics, examining various possibilities for predicting and utilising biomechanical signals/parameters~\cite{Richter.2021}. Among these, i) predicting kinetic data\eg~ground reaction forces, from IMU signals or video images \cite{Johnson.2019, HossainGRF.2023, slijepcevic2024predicting}, ii) predicting kinetic signals\eg~joint contact forces from ground reaction forces and joint angles \cite{Saliba.2018, krondorfer2024predicting}, and iii) forecasting spatiotemporal parameters\eg~stride length, from IMU or smartwatch signals \cite{Baileysmartwatches.2023, Verbieststridelength.2023, Vandermeerensteplength.2020, SharifiRenani.2020}, have garnered significant attention.

\textbf{Walking.} Kinetic signals are usually derived from force platform data, which offer high temporal and spatial resolution but require proper physical contact with the platform. However, there are situations where proper contact may not be possible\eg~when individuals have very small stride lengths and widths or are unable to walk unassisted. In addition, force platforms are often permanently installed in laboratories, which restricts measurements to laboratory environments. The latter limitations have led to efforts to derive kinetic signals from alternative measurement approaches such as IMU or video data. Most studies focused on the estimation of ground reaction forces and joint moments from IMU or video data. These studies reported high correlations to the signals calculated with the use of force platforms and inverse dynamics~\cite{Boswell.2021, Aljaaf.2016, Mundt.2020b}. Different ML methods have also been investigated and superior performances were reported for Neural Networks and Random Forests~\cite{Moghadam.2023, Giarmatzis.2020}.

The reported errors in these estimations may be tolerated due to the ability to conduct measurements in natural environments. Moreover, this flexibility can be particularly beneficial in rehabilitation settings, where patients may have pathologies or walking aids that impede clean measurements of ground reaction forces. In such ambulatory tasks, joint moments can be predicted with acceptable error rates using kinematic data collected from IMUs~\cite{Camargo.2022}. There are more kinematic variables that can be predicted based on three-dimensional optical measurements, such as centre of gravity~\cite{Moon.2022} and foot clearance\ie~the distance of the toes to the ground~\cite{Lai.2012}. These studies are mostly conducted with elderly individuals that have a higher risk of falling or tripping because of an altered walking pattern. The recognition of these walking alterations could bring a huge advance to practical walking analysis because at the moment only one step or gait cycle (on the force platform) can be used to calculate the relevant joint moments.

Biomechanical measurements are not only limited to kinetics and kinematics but can also include muscle activity and muscle forces. These forces can also be predicted by kinematic data~\cite{Dao.2019} or used to predict kinetic data such as joint power~\cite{Heeb.2022}. Additional optical~\cite{Zhang.2022} or sensor driven measurements~\cite{Sun.2022} can help to improve those predictions. Furthermore, combining ML with pressure insole data offers a promising solution for predicting ground reaction forces~\cite{de2023auditory, iber2021mind}. 

The estimation of spatiotemporal parameters from IMU signals or smart watches have been extensively studied in both healthy and pathological populations, considering various sensor locations~\cite{Baileysmartwatches.2023, Verbieststridelength.2023, Vandermeerensteplength.2020, SharifiRenani.2020}. Studies have found that the estimation of temporal walking parameters tends to be more accurate than spatial parameters~\cite{SharifiRenani.2020}. With long-term recordings in natural environments and feature estimation techniques, IMUs offer new possibilities for continuously monitoring spatiotemporal walking parameters. These could serve as valuable tool to complement established clinical walking tests or aid in the early detection of diseases. However, it is important to note that further research is required to determine the practical utility of spatiotemporal gait parameters from feature estimation techniques in these applications. This is particularly important given the varying and reduced accuracy compared to the gold standard.

\textbf{Running.} Feature estimation has been used in running to estimate kinetic variables that are considered risk factors for running-related injuries. This encompasses investigations that predicted ground reaction force~\cite{Johnson.2018, Begg.2006}, joint force~\cite{Johnson.2018}, and bone force~\cite{Matijevich.2020} signals and features from IMU and video data. It is noteworthy that the vertical ground reaction force loading rate could be predicted using only a single IMU sensor~\cite{Derie.2020}. While ongoing advancements hold the potential to improve accuracy and provide valuable tools for injury prevention, it is important to acknowledge that current accuracy levels are often not yet suitable for practical applications.

\textbf{Sports Movements.} Feature estimation has been explored in sports movements, such as cutting, side-stepping, squatting, or jumping. The predicted features included joint moments and forces~\cite{Johnson.2019, Stetter.2019, Chaaban.2021, Zago.2019}, ground reaction forces~\cite{Johnson.2019}, lower-body kinematics~\cite{Chen.2021, Choffin.2022}, and electromyographic variables~\cite{Moniri.2021}. Only a few studies have investigated feature estimation techniques in sports-specific movements. For example, arm kinetics have been used to predict fastball velocity~\cite{Nicholsonfastball.2022} and arm kinetics~\cite{Nicholsonkinetics.2022} in baseball pitching. In handball, the centre of mass kinematics during a jump shot can be accurately predicted based on force platform signals~\cite{Akl.2017}. Additionally, IMU signals have been employed to predict elbow joint angle during swimming with minor errors compared to marker-based camera systems~\cite{Macaro.2018}.

\subsubsection{Limitations and Future Perspectives}
Feature estimation represents a key application of ML approaches in biomechanics, with the aim of enhancing gait and sports biomechanical testing.
However, there are challenges and opportunities for further development.

\textbf{Limited Data.} Particularly evident in sports movements, feature estimation approaches often encounter limitations due to their dependence on relatively small datasets. The potential of data augmentation techniques to mitigate this issue remains largely unexplored, despite evidence of their effectiveness in enhancing other biomechanical tasks~\cite{dindorf2024enhancing}. Additionally, encouraging collaboration among various gait laboratories shows potential for increasing the quantity of available data, which can help improve feature estimation algorithms.

\textbf{Accuracy in Sports.} Despite significant advancements, achieving high accuracy in feature estimation, particularly within the context of sports movements, remains a paramount challenge. This challenge can be attributed to two primary factors: i) Sports movements exhibit a notably elevated level of complexity compared to activities like walking or running. The complexity of movements, varying intensities, and dynamic nature of sports activities impose considerable demands on estimation algorithms. Consequently, predicting complex biomechanical parameters, such as muscle forces, becomes inherently challenging; ii) Another challenge in sports biomechanics is the lack of reference data for training ML models. Unlike the more standardised testing protocols and publicly available datasets in walking and running biomechanics, the diverse and specific nature of sports movements presents obstacles in acquiring data. Consequently, there is typically only limited data accessible for training ML models, particularly for complex parameters like muscle forces.

\textbf{Challenges in Quantifying Biomechanical Parameters.} Unlike measurable parameters like joint angles and ground reaction forces, muscle forces pose a unique challenge in biomechanical analysis due to the absence of direct measurement techniques. This inherent complexity complicates the feature estimation process in walking, running, and sports biomechanics. Muscle forces are rather approximations than precise measurements, which introduces uncertainties into the training data for feature estimation algorithms. Technological advancements such as instrumented implants offer potential solutions for validating these estimations~\cite{Zarghamcontactforces.2019}.

\textbf{Black-Box Models.} In feature estimation tasks within biomechanical research, there is a prevalent emphasis on predictive performance at the expense of model interpretability. Complex ML models often utilised for feature estimation tasks often lack transparency regarding the input features and their relationships driving these predictions. This does not only hinder biomechanists' ability to extract meaningful insights but also undermines trust in the systems.

These limitations present opportunities for advancement. In particular, the integration of physics-informed ML has shown significant potential in enhancing feature estimations, as exemplified by the work of Taneja et al.~\cite{TanejaPINN.2022,zhang2022physics}. physics-informed ML entails the incorporation of pre-existing knowledge concerning the fundamental physical principles into the training process of ML models. Through this integration, the model can more effectively capture the underlying dynamics of the system under investigation, thereby yielding more precise and interpretable predictions.

\subsection{Event Detection}
\label{sec:event}
\subsubsection{Problem Description}
The goal of event detection in time series data is to annotate certain events to extract useful and vital information or remove unwanted and unnecessary data for further analysis. When analysing movement, recurring cyclical movements like running, swimming, cycling, and walking, and non-cyclical movements such as football, handball, golf, and table tennis can generate a vast amount of time series data. Postprocessing the captured data requires a segmentation of the movement based on certain events. A correct segmentation and detection of events is a crucial task due to the calculation of movement and sports-specific metrics based on the segmentation. With these metrics, a person-specific score or movement pattern can be analysed, which can give information about running performance and injuries~\cite{rivadullaDevelopmentValidationFootNet2021}, sports-specific game performance of players\eg~passes, kicks, and sprints ~\cite{komitovaTimeSeriesData2022, kranzingerClassificationHumanMotion2023}, and indicating morbidity, mortality, or risk of falls in walking~\cite{zhangFeelingsDepressionPain2020, patelWalkingRelationMortality2018, brodieEightWeekRemoteMonitoring2015, vergheseMobilityStressTest2012, mDailyWalkingIntensity2012}. Furthermore, supervised ML approaches can be trained on the segmented data to discover hidden relationships in the underlying multidimensional time series data for further investigations and meaningful insights~\cite{komitovaTimeSeriesData2022}. 

\subsubsection{Approaches}
Both walking and running can be segmented into gait cycles. A single gait cycle can be further divided into a stance and a swing phase. The stance phase starts with the foot's initial contact (IC) on the ground and ends with the same foot leaving the ground, known as the foot off (FO). This FO initiates the swing phase, which ends with the IC of the same foot, simultaneously ending also the gait cycle~\cite{bakerMeasuringWalkingHandbook2013}. This segmentation of the gait cycle allows the calculation and further analysis of spatiotemporal, kinematic, and kinetic parameters for each person. The gold standard for detecting IC and FO events in time series data is by defining a certain threshold on the vertical ground reaction force data. However, in gait analysis practice, force platform contacts are often scarce or entirely unavailable due to pathological characteristics or when conducting measurements in the field~\cite{dumphartRobustDeepLearningbased2023}. For studies focused on running, conventional treadmills are commonly preferred over force-instrumented treadmills due to their cost-effectiveness~\cite{rivadullaDevelopmentValidationFootNet2021}.

In cases where no force platforms are available, a manual identification of gait cycle events is often performed. However, depending on various factors, the identification of gait events can be subjective, tedious, and time-consuming work, which is prone to human error~\cite{visscherValidationStandardizationAutomatic2021}. 

Heuristic-based event detection algorithms have been developed to automate the segmentation process. They are able to automatically detect gait events based on local minima and maxima in kinematic data of markers of the lower extremities. However, heuristic algorithms mostly depend on predefined thresholds, which can vary based on various parameters including walking/running speed, foot strike pattern, and marker configuration~\cite{dumphartRobustDeepLearningbased2023, rivadullaDevelopmentValidationFootNet2021}.

Probabilistic-based approaches have been introduced to overcome issues related to heuristic-based algorithms. With recent advancements, ML and Deep Learning methods are able to outperform heuristic algorithms in detecting IC and FO events more robustly and accurately~\cite{rivadullaDevelopmentValidationFootNet2021,dumphartRobustDeepLearningbased2023,lempereurNewDeepLearningbased2020, kidzinskiAutomaticRealtimeGait2019}. In the following, different ML-based approaches for walking, running, and sports movements data will be outlined.

\textbf{Walking.} In walking analysis, an error of below four frames (or 27~ms with 150~Hz capturing frequency) of difference to the ground truth has been used as a common ``accuracy threshold'' for proposed algorithms~\cite{dumphartRobustDeepLearningbased2023, visscherValidationStandardizationAutomatic2021, lempereurNewDeepLearningbased2020}. When this threshold is exceeded, data needs to be interpreted with care, as it can potentially result in differences above the minimal detectable change of $5^{\circ}$ for certain kinematic parameters~\cite{wilkenReliabilityMinimalDetectible2012, geigerMinimalDetectableChange2019}. Due to the limiting factors of heuristic approaches, they often tend to exceed this threshold. ML-based event detection algorithms have shown promising results achieving higher performance and more robust event detection compared to heuristic approaches~\cite{dumphartRobustDeepLearningbased2023,lempereurNewDeepLearningbased2020,kidzinskiAutomaticRealtimeGait2019}.

One of the first ML-based event detection algorithms has been developed by Kidzinski et al.~\cite{kidzinskiAutomaticRealtimeGait2019}. The authors utilised a stacked Long-Short Term Memory architecture and trained the model on mostly children with neurological disorders. Kidzinski et al.~\cite{kidzinskiAutomaticRealtimeGait2019} reported a mean accuracy of 10~ms and 13~ms for IC and FO events, respectively, using a capturing frequency of 120~Hz.

An improved event detection algorithm based on a stacked bi-directional Long-Short Term Memory network has been introduced by Lempereur et al.~\cite{lempereurNewDeepLearningbased2020}. With a similar patient population (n=226) as Kidzinski et al.~\cite{kidzinskiAutomaticRealtimeGait2019}, Lempereur et al.~\cite{lempereurNewDeepLearningbased2020} reported a mean accuracy of 6~ms and 11~ms for IC and FO events, respectively with a capturing frequency of 120~Hz. These results outperformed the most popular heuristic-based event detection algorithms in the literature.

Another bi-direction Long-Short Term Memory architecture approach was developed by Kim et al.~\cite{kimDeeplearningApproachAutomatically2022} utilising data from children with cerebral palsy (n=363). Additionally, the authors compared results for different marker placements and striking patterns (heel, midfoot, and toe). This resulted in a 95\% trimmed-average time error of 8~ms and 11~ms for IC and FO, respectively using a capturing frequency of 150~Hz. Furthermore, their results also showed that the marker setup can influence the performance up to 10\% for different striking patterns and needs to be considered for future approaches. 

Filtjens et al.~\cite{filtjensDatadrivenApproachDetecting2020} utilised a temporal Convolutional Neural Network to detect events during turning and freezing of walking in patients with Parkinson´s disease (data were captured with 100~Hz). The authors reported slightly improved results compared to the Long-Short Term Memory model from Kidzinski et al.~\cite{kidzinskiAutomaticRealtimeGait2019}. However, as this study utilised a dataset comprising only 15 Parkinson's disease patients, the results need to be interpreted with care.  
The most recently published automated event detection algorithm by Dumphart et al.~\cite{dumphartRobustDeepLearningbased2023} trained two separate stacked bi-directional Long-Short Term Memory models for the IC and FO events. The authors utilised the most extensive dataset (n=1,211) containing different pathologies\ie~malrotation deformities of the lower limbs, club foot, infantile cerebral palsy and healthy participants (n=61). Their approach resulted in slightly improved results compared to the model from Lempereur et al.~\cite{lempereurNewDeepLearningbased2020} detecting IC and FO events within 5~ms and 11~ms, respectively, using a capturing frequency of 150~Hz. Furthermore, Dumphart et al.~\cite{dumphartRobustDeepLearningbased2023} investigated the generalisability across laboratories, suggesting that models trained on data from a different laboratory need to be applied with care due to setup variations or differences in capturing frequencies.

For field measurements or monitoring daily activities, most of the current literature focuses on the use of single or multiple IMUs placed on the body to detect events. ML-based event detection algorithms have demonstrated superior performance compared to heuristic methods when applied to IMU data. Single-mounted IMU approaches like waist, shank, or foot-mounted IMUs have become the preferred option to enhance user compliance, as they require fewer sensors to be attached. Algorithms based on waist-worn~\cite{arshadDeepLearningBasedGait2023, arshadGaitEventsPrediction2022}, shank-worn~\cite{romijndersDeepLearningApproach2022}, and foot-worn~\cite{zagoMachineLearningBasedDetermination2021} IMUs show a mean average error of 6~ms to 9~ms for IC and 5~ms to 8~ms for FO detection. 

Systems with multiple IMUs are often used to replicate a laboratory analysis when attempting to predict kinematic and kinetic data during walking and other activities~\cite{dorschkyEstimationGaitKinematics2019}. Lin et al.~\cite{linGaitParametersAnalysis2021} developed a Long Short-Term Memory-based approach to for gait event detection based on five IMUs, which achieved a mean absolute error of 2~ms for IC and 18~ms for FO detection.

\textbf{Running.} In running, the error of determining IC and FO events is highly dependent on running speed due to the generally faster movement. A sensitivity study performed by Rivadulla et al.~\cite{rivadullaDevelopmentValidationFootNet2021} identified changes in the sagittal kinematics of the hip, knee, and ankle at IC and FO, categorising them as either acceptable, reasonable, or unacceptable. A difference lower then $2^{\circ}$ is deemed as acceptable, 2 to $5^{\circ}$ as reasonable but requires consideration and above $5^{\circ}$ as unacceptable~\cite{rivadullaDevelopmentValidationFootNet2021}. Heuristic event detection algorithms for running are based on certain thresholds which can be affected by running speed and foot-strike patterns, limiting the accuracy and robustness of such approaches. Osis et al.~\cite{osisPredictingGroundContact2016a} proposed an ML approach based on Principal Component Analysis to detect IC and FO events during treadmill running using kinematic data acquired from a motion capture system. The results showed that 95\% of the IC and FO events were detected within a 20~ms window with a capturing frequency of 200~Hz for both forefoot and heel-strike patterns (n=186). However, 20~ms can lead to differences of up to $7^{\circ}$ in the knee flexion angle at relative low running speeds~\cite{osisPredictingTimingFoot2014}, leading to unacceptable errors.

Recently, Rivadulla et al.~\cite{rivadullaDevelopmentValidationFootNet2021} presented the most accurate and robust event detection approach for running based on a stacked bi-directional Long-Short Term Memory model. The authors utilised five different datasets~\cite{weirInfluenceProlongedRunning2019, matijevichGroundReactionForce2019, fukuchiPublicDatasetRunning2017} to overcome issues related to differences in marker configuration, running speed, foot-strike patterns, and the limitations arising from relying on data from a single laboratory. These datasets comprised lower limb kinematic data (n=80 runners) that were captured in three different laboratories at different capturing frequencies. This approach outperformed previously published algorithms by detecting IC events with a root-mean-square error of 5~ms and FO events with 6~ms.

IMU-based event detection in running has been researched in and out of the laboratory but mostly with heuristic approaches~\cite{donahueEstimationGaitEvents2023}. Robberechts et al.~\cite{robberechtsPredictingGaitEvents2021} developed an ML-based event detection algorithm with two accelerometers placed on the tibia. Participants were only rearfoot runners (n=93) and measurements were captured in an indoor running track laboratory (30~m). The algorithm detected IC events with a mean average error of 2.2 $\pm$~3.3~ms and FO events within 4.0 $\pm$~4.52~ms ~\cite{robberechtsPredictingGaitEvents2021}. To address the need for measurements beyond laboratory settings, Donahue et al.~\cite{donahueEstimationGaitEvents2023} developed an IMU-based event detection algorithm, which also estimates kinetic parameters. Data (n=16) was captured using IMUs mounted bilaterally on the dorsal aspect of the foot and on the sacrum, force data from an instrumented insole, as well as GPS data from a sports watch. With a bi-directional Long-Short Term Memory architecture, IC events were detected with a root-mean-square error between 11 and 51~ms and FO events between 20 and 53~ms. 

\textbf{Sports Movements.} In comparison to cyclic movements like walking and running, sports movements are mostly non-cyclic movements where ground truth events are typically not as distinguishable as in cyclic movements. With current advancements in IMUs, ML and Deep Learning, automated segmentation of sports movements has become efficient and accurate for sports performance analysis and has become a vital tool for athletes and trainers~\cite{kranzingerClassificationHumanMotion2023, mcgrathUpperBodyActivity2021a, hartSystematicReviewAutomatic2021}. A recent scoping review~\cite{kranzingerClassificationHumanMotion2023} analysed 93 studies regarding segmentation of IMU data in sports movements using ML. Tennis, football, and basketball, as well as workout training/exercises were the most frequently analysed sports movements in the related literature. However, no general conclusion can be formed about weather Deep Learning or traditional ML methods are superior for sports movements segmentation due to a high variety of different tasks analysed.

Pei et al.~\cite{peiEmbedded6axisSensor2017} utilized an IMU embedded in the tennis racket handle to segment tennis strokes into forehand, backhand, and serve, achieving an accuracy of 96\% for segmentation and 98\% for shot detection. In addition to tennis stroke segmentation, Zhao et al.~\cite{zhaoTennisEyeTennisBall2019} estimated the tennis ball speed for each stroke and Whiteside et al.\cite{whitesideMonitoringHittingLoad2017} the hitting load for each stroke for further player performance evaluation.
Alobaid et al.~\cite{alobaidMachineLearningApproach2018} segmented different football activities (shooting, passing, heading, running. and dribbling) using accelerometer data from a single mobile phone with an accuracy of 90\%. Stoeve et al.~\cite{stoeveLaboratoryFieldIMUBased2021} investigated the shot and pass detection in football with two IMU sensors in each shoe with an $F_1$ score of 0.93. 
Acikmese et al~\cite{acikmeseArtificialTrainingExpert2017} developed a basketball training exercise type segmentation\ie~forward–backward dribbling, left–right dribbling, regular dribbling, two hands dribbling, shooting, and layup, algorithm based on a single wrist-mounted IMU with an accuracy of 99.5\%.
Jeong et al.~\cite{jeongPhysicalWorkoutClassification2019} proposed an algorithm for exercise segmentation, employing a single wrist-mounted IMU to differentiate between pull-ups, barbell rows, bench presses, dips, squats, deadlifts, and military presses, achieving an accuracy of up to 96\%.

\subsubsection{Limitations and Future Perspectives}

\textbf{Model Generalisability.} A limitation for ML-based walking and running event detection algorithms is that the proposed algorithms have not been tested on data from different laboratories. Rivadulla et al.~\cite{rivadullaDevelopmentValidationFootNet2021} used data from different laboratories to train their ML algorithm, however, there was no data present from an unseen laboratory. Dumphart et al.~\cite{dumphartRobustDeepLearningbased2023} showed the need for caution when using an algorithm trained on data from another laboratory, as it can lead to errors of up to~10~ms due to differences in capturing frequency and recording setup. These results motivate future research that should focus on developing standardised gait event detection methods that can demonstrate high accuracy and robustness across various laboratories and movement tasks.

\textbf{Sensitivity for Errors.} Moreover, conducting sensitivity analyses is crucial to evaluate the degree to which event detection errors impact the computation of kinematic or spatiotemporal parameters. Rivadulla et al.~\cite{rivadullaDevelopmentValidationFootNet2021} suggested acceptable, reasonable and unacceptable degree thresholds for sagittal knee kinematics at IC and FO that need to be determined for other movements as well.

\textbf{Transfer to Real-World Settings.} IMU-based walking and running event detection algorithms are commonly only evaluated in laboratory settings, which does not give a complete picture of how well these methods work in the real world where the movement is performed~\cite{bensonThisRealLife2022}. Future research should focus on evaluating these approaches also for field measurements beyond the laboratory setting.

\textbf{Limited Data.} In sports movements, event detection algorithms are often trained on a relatively low amount of subjects\ie~mean n=20~\cite{kranzingerClassificationHumanMotion2023, hartSystematicReviewAutomatic2021}. Collaborations between laboratories could increase the number of subjects as well as further quality testing of the algorithms. Furthermore, data augmentation techniques were rarely used or analysed to mitigate part of this problem. Future research should focus on integrating data augmentation techniques due to the benefits reported in other domains.

\textbf{Skill Differences.} The relatively low amount of data could affect the performance of the models in a differently skilled population\eg~amateur, recreational, or professional athletes due to differences in proficiency of  movements~\cite{kranzingerClassificationHumanMotion2023, hartSystematicReviewAutomatic2021}. This needs to be further investigated and could be an opportunity to explore the use of transfer learning in future work.

\subsection{Data Exploration \& Clustering}
\label{sec:cluster}

\subsubsection{Problem Description}
Biomechanical data in both gait and sports biomechanics is characterised by high inter-individual variability, coupled with a frequent lack of data annotations. Hence, employing unsupervised ML methods is crucial for extracting meaningful insights from such data. Cluster analysis, a fundamental component of data exploration, aims to group individuals with similar biomechanical characteristics,, thereby revealing the underlying structure of the data. With the inherent complexity of the data, ML-driven approaches can reveal subtle subgroups that may elude human perception, thus uncovering latent patterns. Clustering aids in the discovery of latent groups within non-annotated datasets but also facilitates the process of data annotation and analysis while assisting in the detection of outliers and erroneous data samples.

\begin{figure*}
\centering
\includegraphics[width=1\linewidth]{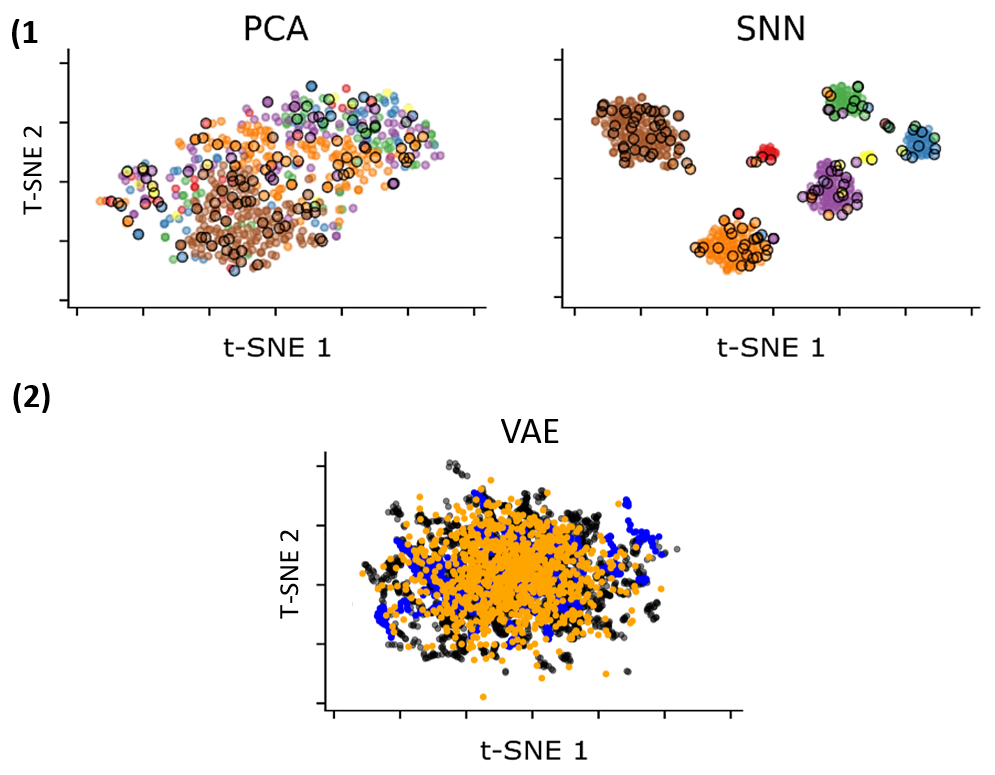}
\caption{Overview of different data exploration approaches for biomechanical data. (1) t-Distributed Stochastic Neighbour Embedding visualisations of principal components (left) and embeddings obtained via metric learning using a Siamese Neural Network (right) for various walking patterns associated with cerebral palsy~\cite{krondorfer2021deep}. (2) t-Distributed Stochastic Neighbour Embedding latent space visualisation of a Variational Autoencoder trained on spinal posture data (black circles correspond to training data, blue to testing data, and orange to sampled latent vectors from the latent space of the trained Variational Autoencoder for data generation \cite{Dindorf.2022vertebrae}).}
\label{fig:Visualizations}
\end{figure*}

In biomechanical applications, clustering based on ML techniques is closely associated with the use of visualisations for effective data exploration. This encompasses tasks such as visually identifying clusters within the data~\cite{Leporace.2021, Dindorf_visualization.2022, krondorfer2021deep}\eg~Figure~\ref{fig:Visualizations}, evaluating class separations in classification tasks~\cite{Liextraction.2022}, and examining different feature extraction methods~\cite{xu2022explaining, Liextraction.2022}. Consequently, methods for dimension reduction and visualisation play a pivotal role in presenting high-dimensional biomechanical data in a lower-dimensional space. 

\subsubsection{Approaches}
\textbf{Walking.} Less complex methods such as Principal Component Analysis or Multidimensional Scaling have been commonly applied in biomedical data exploration to uncover latent patterns~\cite{Hartebrodtbiomed.2022, Uguzpca.2012}. Approaches using Self-organising Maps were also leveraged in the context of walking data and date back to the 1990s~\cite{KohleMaps.1996}. More advanced methods focusing primarily on visualisation as well as learning of latent embeddings were mainly applied to walking data. For visualisation, t-Distributed Stochastic Neighbour Embedding~\cite{Hajian_tsne.2019, LaFuente.2023, slijepcevic2024individual} and Uniform Manifold Approximation and Projection~\cite{LaFuente.2023} were most often used in literature. 
For learning a latent embedding, which can be used for visualisation purposes, but also for preparatory steps in the context of feature extraction, studies used\eg~Autoencoder~\cite{Hernandez_autoencoder.2020} or Siamese Neural Networks~\cite{Dindorf_visualization.2022}. An improvement of separability among subgroups has been observed when t-Distributed Stochastic Neighbour Embedding is applied to learned embeddings, as opposed to its direct application to the raw data~\cite{krondorfer2021deep}.

Clustering has been utilised across various applications aimed at enhancing research and practice in walking biomechanics.
Most research is conducted on patients with different kinds of pathologies, with orthopaedic problems being the most prevalent. Promising results are observed, particularly when clustering patients with arthritic diseases, for example, distinguishing between symptomatic and asymptomatic patients based on biomechanical parameters~\cite{Leporace.2021,YoungShand.2023,Halim.2022,Su.2001}. Different clusters based on sex and gait kinematics can be observed in people with total knee arthroplasty and candidates for this kind of surgery~\cite{YoungShand.2023}. Halim et al.~\cite{Halim.2022} utilised clustering algorithms to identify severity grades in osteoarthitis based on pain scale, kinematic, and electromyographic data from patients with osteoarthritis. Leoprace et al.~\cite{Leporace.2021} identified four clusters based on kinematic variables for walking data from patients with osteoarthritis. Furthermore, chronic instabilities which are hard to diagnose can be differentiated efficiently with clustering methods based on kinematics~\cite{Qian.2020}. In prosthetic research and practice it is not easy to evaluate how well the prosthesis performs. Clustering has been employed to categorise  patients with amputations into homogeneous clusters based on spatiotemporal parameters, revealing distinct patterns within this population~\cite{Ichimura.2022}, which might help to improve the individualisation of prosthetics. Besides orthopaedic diseases, also neurological pathologies\eg~strokes show altered kinematics that have been clustered in the literature~\cite{Kaczmarczyk.2009}.

\textbf{Running.} In running, it is important to detect movements that can lead to injury and optimise the economy of the movement to save energy for longer runs. Hoerzer et al.~\cite{HoerzerMaps.2015} used Self-organising Maps to define functional groups with different movement patterns in running. Pardo Albiach et al.~\cite{PardoAlbiach.2021} evaluated various spatiotemporal parameters and the power output of runners across different running speeds using hierarchical cluster analysis.
Further biomechanical variables such as ground reaction forces, bone kinematics, and electromyographic data have also been clustered to identify different types of runners~\cite{slijepcevic2024individual,Aljohani.2020,Senevirathna.2023,Jauhiainen.2020}. A study by Aljohani et al.~\cite{Aljohani.2020} made use of various biomechanical variables to train a clustering algorithm that identified four different clusters in runners independent from sex. Senevirathna et al.~\cite{Senevirathna.2023} identified two clusters in runners based on kinetic and spatiotemporal parameters.

\textbf{Sports Movements.} Principal Component Analysis and Multidimensional Scaling are also frequently employed in the field of sports biomechanics to uncover hidden patterns in the data~\cite{Deluzio_waveform.2007, Welch_cutting.2021}. Self-organising Maps were also used for kinematic data\eg~for analysing volleyball spike kinematics~\cite{SerrienMaps.2017}. However, in comparison to the analysis of walking and running data, the application of more sophisticated data exploration methods in the field of sports biomechanics is still relatively scarce.

In sports-related movements, clustering is employed to differentiate movement executions from a range of general exercises~\cite{Remedios.2020}. Another important application is the detection of pathologies~\cite{Phan.2022} or movements that are linked to injury~\cite{LaFuente.2023}. Through the identification of particular movement patterns or functional groups within the data, these algorithms have the potential to enhance training programs, particularly for injury prevention and rehabilitation.

Clustering is used in the analysis of sports biomechanics in various kinds of sports. Often it is used to assess the outcome of a special event\eg~volleyball spike based on upper limb angular velocities~\cite{Sarvestan.2021} or serves in tennis~\cite{Kovalchik.2018}. There are also some applications in individual sports such as the clustering of the canoeing motion from different levels of proficiency based on kinematic data~\cite{Liu.2020} or identifying the phenotype of Olympic wrestlers based on anthropometric data~\cite{Bonilla.2022}. Giles et al.~\cite{Giles.2023} analysed a large set of spatiotemporal data from tennis matches to identify patterns in the change of players' movement directions.

Clustering has also be used to evaluate coordination patterns in sports movements\eg~during change of direction using kinematic data~\cite{SarvestanClustering.2022}. Whiteside et al.~\cite{WhitesideCharacteristics.2017} used clustering to evaluate the best position for executing a tennis serve with the goal of achieving an ace. Furthermore, clustering has been utilised to analyse stability of hold during shooting\eg~in biathlon shooting where differences in stability of hold become apparent among athletes with varying levels of performance~\cite{BacaBiathlon.2012}.

\subsubsection{Limitations and Future Perspectives}
Although cluster analysis hold substantial potentials in gait and sports biomechanics, several limitations and challenges exist.

\textbf{Noisy Character of Biomechanical Data.} Biomechanical data is often prone to noise or incomplete data, which can significantly impact the performance and reliability of clustering algorithms.

\textbf{Complexity of Clustering Time Series Data.} Biomechanical data is often collected over time, resulting in time series data. Clustering time series data requires specialised techniques that account for temporal dependencies and variations in movement patterns. Choosing appropriate features from the biomechanical data that capture relevant information for clustering is substantial but also challenging. Furthermore, the high dimensionality of the data can lead to computational inefficiency and may cause some clustering algorithms to perform poorly.

\textbf{Classification of Clusters in Biomechanical Data.} Cluster analysis can divide the dataset into subsets that share similar structural characteristics. This approach can also be viewed as a preprocessing step for\eg subsequent classification tasks. For example, training and evaluating different classifiers on each cluster separately might help to identify the best model for each subgroup of data, leading to more accurate predictions.

\textbf{Labelling of Biomechanical Data.} Cluster analysis can aid in semi-supervised learning approaches by assigning cluster annotations to non-annotated (unlabelled) data. This can provide additional training data for the classifier, leading to better performance, especially in scenarios where annotated data is scarce (see also Section~\ref{subsec:datascarcity}). Cluster analyses could therefore also be of great importance in addressing a central challenge of limited data in gait and sports biomechanics research.

\subsection{Automated Classification}
\label{sec:classification}
\subsubsection{Problem Description}

The analysis of biomechanical data poses challenges owing to its multidimensional and interrelated nature~\cite{chau.2001}. To address these challenges, the application of ML approaches for automated classification has been proposed as a tool to enhance the analysis of biomechanical data. Automated classification refers to the task of predicting the category or class to which a biomechanical data sample belongs. Supervised ML methods, including Support Vector Machines, Random Forests, and Convolutional Neural Networks, have been utilised to conduct automated classifications. Supervised ML methods are employed to train classification models based on a collection of samples, known as training data. Each sample within this training dataset comprises specific input features\eg~3D ground reaction forces and/or 3D marker trajectories, accompanied by corresponding target labels that denote the respective category or class of the sample\eg~healthy or pathological gait. ML methods model the training data and learn relationships between the input features and the corresponding target labels.
Once ML models are trained, they can be utilised for the automated classification of unseen data samples. By employing these ML models, users can leverage data-driven insights derived from extensive datasets.
The literature on the use of supervised ML approaches for the automated classification of biomechanical data is outlined in Section~\ref{subsec:classification-approaches}.

\subsubsection{Approaches}
\label{subsec:classification-approaches}

The ability of supervised ML to automatically classify biomechanical data is illustrated by the following examples, separately for the domain of walking, running, and sports movements.

\textbf{Walking.} In the context of fundamental research on the use of ML for the automated classification in biomechanics, biomechanical data were frequently gathered in laboratory-based settings. In this context, supervised ML methods were employed to test whether biomechanical input features can be classified according to personal or demographic target labels. For instance, supervised ML methods were used to demonstrate that individuals can be accurately identified based on kinetic\eg~99.3\% in 671 individuals~\cite{horst2023modeling}, kinematic\eg~100.0\% in 57 individuals~\cite{horst2019explaining}, and electromyographic\eg~98.9\% in 79 individuals~\cite{hug.2019} input features measured during walking. Further studies showed that biomechanical walking patterns allow for identifying individuals across different days~\cite{horst.2016,hug.2019} and months~\cite{horst.2017}. The identification of individuals based on their walking patterns can find its applications in personalising therapeutic diagnoses and treatments as well as biometrics and security~\cite{connor.2018}.
Besides personal characteristics, supervised ML methods were used to predict demographic characteristics such age~\cite{begg.2005,zhou.2020} and sex/gender~\cite{yoo.2005,pathan.2021} based on biomechanical walking data. 

In the realm of clinical gait analysis, typically conducted in laboratory-based settings, studies have primarily focused on distinguishing between pathological and healthy biomechanics for diagnostic purposes~\cite{figueiredo.2018}. Notable examples include distinguishing patterns in specific conditions, such as stroke~\cite{lau.2009}, Parkinson's disease~\cite{wahid.2015}, cerebral palsy~\cite{alaqtash.2011,slijepcevic.2023}, multiple sclerosis~\cite{alaqtash.2011}, osteoarthritis~\cite{nuesch.2012}, anterior cruciate ligament injury~\cite{christian.2016}, and functional walking disorders~\cite{slijepcevic.2017}. The latter studies demonstrate the effective utilisation of ML models for classifying patient groups. These findings suggest that ML models can be developed as decision support tools for diagnostic purposes, aiding clinicians in making informed decisions based on biomechanical data. 

Supervised ML methods also demonstrated their potential in predicting fall risk among the elderly~\cite{lockhart.2021} and patients with neurological disease~\cite{filtjens.2021,shalin.2021}, contributing to preventive strategies. By analysing long-term data in field-based settings, these models can forecast the likelihood of falls and empower proactive interventions.

\textbf{Running.} In laboratory-based settings, supervised ML approaches for automated classification of biomechanical data during running were used for the classification of individual runners~\cite{hoitz2021individuality}, age~\cite{fukuchi.2011,phinyomark.2014}, sex/gender~\cite{phinyomark.2014,maurer.2012}, running speed~\cite{maurer.2012}, and running shoe~\cite{maurer.2012,trudeau.2015}.

A trend that can be observed in recent years is that supervised ML approaches predominantly utilised biomechanical data acquired from wearable IMU sensors in both laboratory- and field-based settings. This trend is linked to the widespread accessibility of such sensors on the consumer market and their extensive utilisation for tracking movements in everyday life and running across performance levels. Supervised ML approaches were able to classify wearable sensor data during running according to different types of outdoor terrains~\cite{dixon.2019}, inclinations of surface~\cite{ahamed.2019}, environmental weather conditions~\cite{ahamed.2018}, levels of fatigue~\cite{marotta.2021}, subjective perceptions of running shoe comfort~\cite{koska.2020}, and runner's performance level~\cite{clermont.2019,liu.2020performance}.

\textbf{Sports Movements.} In laboratory-based settings, supervised ML methods were also implemented to accurately identify individuals based on their specific movement patterns in pedalling~\cite{hug.2019,aeles.2021}, karate~\cite{burdack.2020fatigue} and throwing~\cite{horst.2020}. Burdack et al.~\cite{burdack.2020fatigue} also demonstrated that changes in intra-subject movement patterns were detectable during a karate kicking task, suggesting that adaptations in individual movement patterns occur during the process of fatigue-related accumulation. Novatchkov and Baca~\cite{novatchkov.2013} explore how supervised ML methods can be used for the automated classification of execution quality in weight training. 

In the sports domain, wearable IMU sensors have also been used in field-based settings for activity classification in various sports\eg~in skateboarding~\cite{groh.2015}, snowboarding~\cite{groh.2016}, beach volleyball~\cite{kautz.2017}, and trampoline gymnastics~\cite{woltmann.2023}.

\subsubsection{Limitations and Future Perspectives}

Supervised ML approaches have shown promising performance in automated classifications of biomechanical data on various classification tasks, as highlighted in Section~\ref{subsec:classification-approaches}. Despite promising performances in research, the practical implementation of ML models for automated classification of biomechanical data in clinical settings and athletic routines remains limited. This is related to the fact that many of the ML approaches described in the literature tend to be pilot studies in nature and suffer from the following limitations.

\textbf{Biased Performance Evaluation Regimes.} Shortcomings in the evaluation of ML models are commonly observed in the literature. This includes that several studies distributed the available samples randomly across cross-validation folds in k-fold cross-validation configurations, without controlling an equal distribution of classes and confounding factors in each fold and without controlling that the data of one subject is either part of train, validation, or test data. Moreover, differences in the number of individuals between categories/classes\ie~unbalanced datasets were not taken into account in the performance evaluation of the ML models.

\textbf{Over-Simplified Classification Task.} Studies have primarily used rather simple\ie~binary classification tasks\eg~healthy control vs. Parkinson's disease. For evaluating the general suitability of ML algorithms for automated classification of biomechanical data, a simplification of the classification task\eg~to a binary task with two clearly distinct classes is a reasonable research strategy. However, ML models in clinical practice will need to be able to solve advanced tasks, such as early detection, differential stage classification, and decision-support for medication adjustment in Parkinson's disease. 

\textbf{Class-Dependence.} During model training, we might have only a limited number of specific classes to work with. However, in real-world applications, the model is likely to encounter additional classes that weren't part of its training dataset. Therefore, unsupervised outlier detection methods have recently been proposed which can function as class- or pathology-independent classifiers~\cite{Dindorf.2021spinal, Teufl.2021detection} demonstrating that this approach is preferable to a two- or multi-class classifier in the described scenario~\cite{Hempstalk}.

\textbf{Black-Box Models.} Studies often prioritise predictive performance over interpretability of ML models. While complex, non-linear ML models can accurately classify or predict biomechanical outcomes, the underlying factors and relationships contributing to these predictions may remain opaque. This lack of interpretability hampers the ability of biomechanists to gain meaningful insights and fully understand the mechanisms driving the predictions of the ML models. A fact that may also reduce practitioners' trust in ML models for automated classification of biomechanical data.

The cornerstone for future development of ML models for automated classification is the availability of large-scale, high-quality, and multi-modal benchmark dataset from multiple laboratory- and field-based settings for given classification task (see also Section~\ref{subsec:datascarcity}). In addition to ongoing research and development aimed at enhancing the performance, robustness, and interpretability of ML models for automatic classification in biomechanics, it is imperative to assess their practical applicability. This entails evaluating whether clinical diagnosis can be enhanced with the assistance of ML models, whether accurate and continuous recording of the ground surface, fatigue, and gradient during running using wearable IMU sensors can effectively improve the performance of runners, and whether automatic classifications of fall risks can indeed lead to a reduction in the number of falls among elderly individuals or patients with neurological diseases in everyday life.

\section{Key Limitations of Machine Learning in Biomechanics}
\label{sec:limitations}

In this section, we will address key limitations of ML in biomechanics that are applicable across the described key applications, including explainability~ (Section~\ref{subsec:explainability}) and 
data scarcity~(Section~\ref{subsec:datascarcity}). Further limitations that pertain primarily to individual applications will be discussed in Section~\ref{subsec:furtherlimitations}.

\subsection{Explainability}
\label{subsec:explainability}

\subsubsection{Problem description}

In various gait data analysis tasks, researchers frequently employ white box models like Linear Regression or Decision Trees due to their transparent decision-making processes. In Linear Regression, the relationship between input variables and the output is explicitly defined through coefficients. Similarly, Decision Trees offer a clear decision-making pathway based on a series of if-else conditions, enhancing the interpretability of the model. These white box models enable researchers to understand the relationships between input features and the predicted outcome effectively. However, as ML approaches in biomechanics evolve, there is a growing trend towards utilising complex, non-linear Deep Learning models such as Convolutional Neural Networks. Despite their promising performance in classifying walking, running, and sports movements, these models pose a notable challenge due to their opaque, black-box nature. The lack of transparency presents a challenge in understanding how these models arrive at specific decisions and which underlying patterns and rules they learn from the data.

Explainable Artificial Intelligence (XAI), a field that encompasses different explainability approaches to shed light on the inner workings of complex, non-linear ML models, has gained increasing attention in recent years as an answer to this challenge. In the following, we outline the motivations behind the utilisation of explainability methods in biomechanics. The primary emphasis is on improving the transparency and understanding of the decision-making processes of ML approaches for domain experts, with particular significance in clinical applications. In this context, the critical aspect is to increase trust and acceptance among clinicians and patients into ML-based systems. Explainability methods should, therefore, enable clinical experts to verify decisions and eliminate potential biases by revealing which features they use for predictions. 
Furthermore, ensuring legal and regulatory compliance is another important aspect. Given the stringent regulations that have been implemented in recent years, especially in relation to the use of automated predictions, particularly within the healthcare sector\eg~the EU General Data Protection Regulation~\cite{regulation2016regulation} and the proposed AI Act\footnote{\url{https://europa.eu/!PH46HB} (accessed on 28.09.2023)}. Another motivating factor is the desire for personalised medicine. Due to the highly individualised nature of human movement, decision support and treatment planning often need to be customised to meet each patient's specific needs. Explainability methods could offer in-depth insights into the individual movement patterns responsible for particular decisions. The analysis of individual movement patterns is also relevant for running and sports applications. For example, identifying the individual characteristics within biomechanical data concerning athletes and their sports equipment can aid in performance assessment and improvement, the monitoring of training progress, and the reduction of injury risks.
Finally, explainability methods enable ML engineers to deeper understand the utilised ML methods. This understanding facilitates the evaluation of the effects of data preprocessing procedures and different ML methods. This evaluation considers the classification strategies and utilised input features, extending beyond mere performance assessments. Explainability methods can also help in identifying and mitigating biases, thus optimising ML approaches to align with the specific requirements and tasks within the realm of biomechanics.

\begin{tcolorbox}[width=\textwidth,
                  colback=gray!10,
                  colframe=black,
                  boxrule=1pt, 
                  arc=5pt,
                  boxsep=5pt,
                  left=10pt,
                  right=5pt,
                  top=5pt]
  \textbf{Key Limitation: Explainability}
  
  Explainability in this context entails providing a human-understandable elucidation of the internal mechanisms and decision-making processes inherent to complex non-linear ML models utilised in biomechanics.
\end{tcolorbox}

\subsubsection{Approaches}
According to the taxonomy of Arya et al.~\cite{arya2019one}, three different XAI approaches can be distinguished: data exploration, decision explanation, and model explanation. 

\textbf{Data Exploration.} approaches do not explain the ML model, but rather the data that is used to train the model. The goal of these methods is to uncover meaningful data structures and patterns and to help domain experts understand the data distributions within the data. Various methods have been employed for data exploration\eg~inferential statistical tests, such as Statistical Parametric Mapping~\cite{pataky2010generalized} for time series data~\cite{slijepcevic2021explaining}, visually inspecting data (average waveforms, boxplots, and feature space examination)~\cite{slijepcevic.2017,hausmann2021measuring}, as well as more advanced data exploration techniques from the field of visual analytics~\cite{wagner2018kavagait}.

\textbf{Decision Explanations.} provide \textit{local} explanations for predictions on individual data samples. These explanations identify which input features contribute to a certain prediction. The majority of these approaches are applied in a post-hoc manner to previously trained ML models.

\textbf{Model Explanations.} explain the \textit{global} operation of a trained model and allow to reveal task- and class-specific prototypes and characteristic patterns. These explanations allow domain experts to identify ambiguous features, detect overlap between classes, and assess the correctness of the model.

\textbf{Walking.} Decision and model explanations have been explored using propagation- and perturbation-based explainability methods for walking data. Slijepcevic et al.~\cite{slijepcevic2021explaining} utilised Layer-Wise Relevance Propagation , a propagation-based explainability method, to obtain class- and model explanations for the automated classification of functional walking disorders. The authors proposed a model explanation approach based on averaging Layer-Wise Relevance Propagation relevance scores across all samples within a class, alongside an alternative approach relying on SpRAy~\cite{lapuschkin2019unmasking}, a method for identifying clusters among explanations. 
The explainability results highlighted that the investigated ML models (Support Vector Machines, Multilayer Perceptrons, and Convolutional Neural Networks) base their predictions on features that align with a statistical and clinical evaluation. In addition, the explainability analysis enabled the identification and comparison of learning strategies employed by the different ML methods. Moreover, the explainability results also confirmed the significance of data normalisation for obtaining unbiased predictions from ML models (addressing bias introduced by different signal amplitudes).

Layer-Wise Relevance Propagation has been further explored to provide insights into sex- and age-dependent walking patterns learned from ML models~\cite{horst2020explaining, slijepcevic2022explaining}. Several studies have utilised Layer-Wise Relevance Propagation to identify the features used by ML models for classifying individuals based on their biomechanical walking patterns~\cite{horst2023modeling, horst2019explaining, aeles.2021, slijepcevic.2023signature}.
The explainability analysis showed that unique person-specific walking characteristics, so-called gait signatures, can be determined very reliably~\cite{slijepcevic.2023signature}.

Rind et al.~\cite{rind2022trustworthy} introduced \textit{gaitXplorer}, a visual analytics approach designed for the classification of walking patterns associated with cerebral palsy. The authors employed the propagation-based method Grad-CAM~\cite{selvaraju2017grad} to provide decision explanations for Convolutional Neural Networks. Slijepcevic et al.~\cite{slijepcevic2023explainable} proposed model explanations by aggregating individual Grad-CAM explanations (using the median) for classifying walking patterns associated with cerebral palsy. Additionally, the authors compared these explanations with feature importance from tree-based models\ie~Decision Trees and Random Forests.

Dindorf et al.~\cite{dindorf2020interpretability} utilised the perturbation-based explainability method Local Interpretable Model-Agnostic Explanation~\cite{ribeiro2016should} to explain linear Support Vector Machines trained on kinematic and kinetic waveforms as well as discrete features to differentiate between healthy individuals and patients who underwent total hip arthroplasty. To derive global model explanations, the authors averaged the decision explanations per class. {\"O}zates et al.~\cite{ozates2021application} employed Local Interpretable Model-Agnostic Explanation to explore the discrete features that ML models relied on for distinguishing among nine different foot conditions. 

Kokkotis et al.~\cite{kokkotis2022leveraging} leveraged the perturbation-based explainability method Shapley Additive Explanations~\cite{lundberg2017unified}. The authors applied this method to explain the most influential discrete kinematic and kinetic features for classifying patients with anterior cruciate ligament injuries (both with and without reconstruction surgery) and healthy controls. The authors observed that Shapley Additive Explanations emphasised several discrete features consistent with biomechanical findings in the literature. However, they noted a disparity between the explainability results and the outcomes of traditional statistical analyses.

Furthermore, Dindorf et al.~\cite{dindorf2023machine} investigated Counterfactual Explanations for the task of identifying the presence of hyperlordosis and hyperkyphosis based on postural data. The authors utilised handcrafted domain-specific features and a Gaussian process classifier. 

\textbf{Running.} Only a limited number of studies has utilised explainability methods within the domain of running biomechanics. The identified studies all utilised the propagation-based approach Layer-Wise Relevance Propagation. 
Hoitz et al. utilised Layer-Wise Relevance Propagation to identify running signatures from kinematic and kinetic data from 50 recreational~\cite{hoitz2021individuality} and 20 highly trained runners~\cite{hoitz2021isolating}.
Xu et al.~\cite{xu2022explaining} utilised Layer-Wise Relevance Propagation to identify differences in running patterns between high- and low-mileage runners based on kinematic and kinetic data. Furthermore, Horst et al.~\cite{horst2023identification} leveraged Layer-Wise Relevance Propagation to identifying running signatures of 30 healthy subjects and evaluate the influence of four footwear conditions on individual running characteristics.

\textbf{Sports Movements.} Concerning perturbation-based methods, Lisca et al.~\cite{lisca2021less} employed Shapley Additive Explanations to explain an Extreme Gradient Boost~\cite{chen2016xgboost} classifier trained for the classification of goalkeeper movements in football based on IMU-derived kinematics of the goalkeeper's hand.
Woltmann et al.~\cite{woltmann2023sensor} utilised Shapley Additive Explanations to provide explanations for the prediction of various ML models trained for classifying jumps in trampoline gymnastics. For this purpose, classification relied on time discrete features extracted from the kinetic data of an IMU sensor. 

\subsubsection{Limitations and Future Perspectives}

The utilisation of explainability methods in biomechanics is still relatively in its infancy. Although there is a growing interest in the use of explainability methods in the domain of walking biomechanics, their application in the domain of running, and particularly in the biomechanics of sports movements, remains comparatively unexplored.
Initial research, primarily centred on human walking, suggests considerable potential for advancing our understanding of ML in biomechanics and its capacity to enhance ML approaches. For example, explainability methods allow to understand the functioning of ML models and to show which input features were used for their predictions. In addition to performance-based comparisons, explainability methods allow to identify and compare the classification strategies of different ML methods and the effects of different data preprocessing procedures~\cite{slijepcevic2021explaining, horst2019explaining}. This can aid in detecting biases in ML models and help ML engineers to mitigate or minimise these biases.

A fundamental challenge in evaluating explainability results is the absence of a ground truth. To address this challenge, Slijepcevic et al.~\cite{slijepcevic2021explaining} proposed a two-step approach, using Statistical Parametric Mapping and clinical domain experts, for evaluating the explanations. However, there is a need for more extensive examination of different explainability methods to ensure their meaningfulness and validity. At present, different explainability methods have been utilised\eg~Layer-Wise Relevance Propagation, Grad-CAM, Local Interpretable Model-Agnostic Explanation, Shapley Additive Explanations, and Counterfactual Explanations, however, it remains ambiguous which explainability method is most appropriate for certain biomechanical data and tasks. Thus, similar to the lack of a systematic assessment for the elements of the general ML pipeline, there is also a noticeable gap in methodically evaluating various explainability methods for interpreting biomechanical data. A systematic comparison of explainability methods for different biomechanical tasks and data could contribute to making these methods an essential part of ML approaches within the community. These systematic comparisons should also investigate the possibilities of trainable explanation mechanisms incorporated
directly into complex Deep Learning models such as the method bounded logit attention~\cite{baumhauer2022bounded}. 

An essential consideration for the usefulness of explainability methods lies in determining the specific form\eg~interactive counterfactuals in which domain experts require explanations to ensure their practical value. Slijepcevic et al.~\cite{slijepcevic2021explaining} illustrated that ML models often learn an over-complete set of relevant features, and these features may not always align with all clinically relevant parameters~\cite{slijepcevic.2023}. This discrepancy could potentially diminish experts' trust in the utilised ML methods, even though these methods are effective. Empirical studies assessing the usefulness, trustworthiness, and other aspects of explanations in biomechanics are currently lacking and should be addressed in future research. 

\subsection{Data and Annotation Availability}
\label{subsec:datascarcity}

\subsubsection{Problem Description}

A prevalent limitation of ML approaches in biomechanics is the reliance on small datasets. ML approaches have been predominantly trained and evaluated on data from a limited amount of individuals, gathered within a single laboratory- or field-based setting. This can be attributed to various factors, including challenges in participant recruitment, the need for specialised expertise, and the expensive and complex nature of measurement recording during biomechanical testing. 
Further limitations emerge due to the inability to record data from entire populations, which can introduce various biases into the datasets. These biases might manifest as imbalanced distributions of sex, age, body mass, body height, speed, and performance levels among the categories of a dataset.

The lack of available large-scale benchmark datasets also hinders the comparability and reproducibility of results across studies and limits the widespread adoption of ML-based approaches in biomechanics. The existing literature has a notable lack of comparable classification tasks and datasets. This results in insufficient guidance on appropriate data preprocessing and classification pipelines, including\eg~feature engineering, scaling, and ML methods, suitable for given classification tasks. 

A significant limitation for biomechanical data is also the lack of annotations, which are especially necessary in a clinical setting\eg~pathological gait patterns. Annotations typically require a subjective assessment of the biomechanical data by domain experts. Thus, determining annotations is a labour-intensive and time-consuming task.

\begin{tcolorbox}[width=\textwidth,
                  colback=gray!10,
                  colframe=black,
                  boxrule=1pt, 
                  arc=5pt,
                  boxsep=5pt,
                  left=10pt,
                  right=5pt,
                  top=5pt]
  \textbf{Key Limitation: Data and Annotation Availability}
  
  Data and annotation availability refers to the limited availability or insufficiency of relevant and comprehensive datasets, task-specific annotations for conducting thorough analyses and research in biomechanics.
\end{tcolorbox}

\subsubsection{Approaches}

\textbf{Walking.} Public datasets are already accessible for the purpose of data analysis, and a recent systematic review by David et al.~\cite{david2023human} presents a overview of existing datasets. Most of these datasets contain a relatively small number of individuals and walking trials conducted under various settings and using different measurement devices, as documented in previous studies~\cite{Moore.2015, LuoYue.2020, Fukuchi.2018, Lencioni.2019, Schreiber.2019, CamaraMiraldo.2020}. However, there are a few datasets that offer more potential for ML approaches, such as the GaitRec dataset~\cite{horsak.2020}, Gutenberg Gait Database~\cite{horst.2021}, and a dataset published by Derlatka and Parfieniuk~\cite{derlatka2023real}. The GaitRec dataset is one of the most comprehensive datasets in the field containing bilateral ground reaction force data of 2,084 patients and 211 healthy controls with a total of 75,732 walking trials. The GaitRec dataset contains clinically relevant annotations determined by a clinical expert. These annotations specify the general anatomical joint level at which an orthopaedic impairment is located, such as at the hip, knee, ankle, or calcaneus, and the dataset provides also more detailed annotations about the impairment type.
The Gutenberg Gait Database contains also bilateral ground reaction force data of 350 healthy controls with a total of 8,819 walking trials. The dataset by Derlatka and Parfieniuk~\cite{derlatka2023real} comprises 13,702 walking trials of bilateral ground reaction force data from 324 healthy controls wearing shoes of various types. 

Regarding the standardisation of data preprocessing, the authors of the GaitRec dataset proposed a general preprocessing pipeline for ground reaction force data including thresholding, filtering, outlier detection, and an export format. This pipeline was also employed for the Gutenberg Gait Database in order to enable the seamless utilisation of both datasets.

In the ML literature, various strategies have been introduced to address the challenge of dealing with limited data or sparsely labelled data. Few-shot, including variants such as one- and zero-shot learning approaches have already been explored for the analysis of walking data~\cite{krondorfer2021deep, Dindorf.2022vertebrae, duncanson2023deep}. These approaches have the advantage of being able to model effectively even limited amounts of data per class. This means that they can extract information from only a few samples per class and create models that are able to detect and classify previously unseen classes. This is especially important in situations where data collection is limited as with biomechanical data\eg~due to rare pathological walking patterns. 

Inter-subject variability is another challenge that can negatively impact model quality. To deal with this challenge, Dindorf et al.~\cite{Dindorf.2022vertebrae} were able to achieve the first promising results using Siamese Neural Networks as automated feature extractor for learning features with reduced intra-subject variability between different walking cycles or measurement days.

\textbf{Running.} In the context of running, there are also publicly accessible datasets available~\cite{david2023human}. However, the number of individuals in these datasets is limited compared to the aforementioned walking datasets. A common characteristics of running datasets is the substantial volume of data gathered for these limited number of individuals, given the recording of numerous running cycles. Running data is frequently a subset of a more extensive dataset that includes both everyday activities, in the context of human activity recognition, as well as specific sports-related movements. The available running data are commonly recorded in a laboratory-based setting via a marker-based infrared camera systems~\cite{rivadullaDevelopmentValidationFootNet2021, matijevichGroundReactionForce2019, fukuchiPublicDatasetRunning2017, estevez2015open} or IMU-based sensors~\cite{khandelwal2017evaluation, chereshnev2018hugadb} and an instrumented treadmill. The biomechanical assessment also extends beyond the laboratory setting and into the field through the use of IMU-based sensors~\cite{khandelwal2017evaluation}.

\textbf{Sports Movements.} Data availability presents an even more pronounced challenge in sports movements compared to walking and running. Collecting comprehensive biomechanical data on sports movements can be challenging and resource-intensive. Moreover, in the realm of high-performance sports , it is common to encounter scenarios in which data collection is restricted to a highly exclusive group of individuals, usually professional athletes. Their data is usually kept confidential to maintain a competitive advantage, among other reasons.

Similar to running data, biomechanical signals extracted from sports movements often form a subset within the broader scope of datasets used in the domain of human activity recognition. These datasets\eg~HuGaDB dataset~\cite{chereshnev2018hugadb}, MOCAP-ULL Database~\cite{estevez2015open}, and SFU Motion Capture Database\footnote{\url{https://mocap.cs.sfu.ca/}~(accessed on 28.09.2023)}~ cover a wide range of movements, including activities like jumping, parkour rolls over objects, cartwheels, bicycling, martial arts kicking, and dance sequences. One of the most extensive datasets that includes sports movements is the MoVi dataset~\cite{ghorbani2021movi} that encompasses records from 90 actors engaging in a total of 21 diverse activities, including both everyday actions and sports -related movements\eg~crawling, vertical jumping, jumping jacks, kicking, throwing, and catching.

\subsubsection{Limitations and Future Perspectives}

Data availability is a major challenge for ML approaches in biomechanics. To date, datasets have been made publicly available to varying degrees, but there is a clear imbalance between the availability of datasets in the walking, running, and sports movement domain. A relatively large number of datasets have been published regarding human walking, while the availability of datasets related to running and sports movements remains comparatively limited. Methods for learning with limited data\eg~zero-shot learning, have been employed for walking data, but their potential applications in the field of running and sports movements hold promising prospects for the future. 

There exists a high demand for high-quality, large-scale benchmark datasets collected in diverse laboratory and field settings. Benchmark datasets would allow for systematic comparisons of various ML methods, as exemplified by recent studies~\cite{burdack.2020,slijepcevic.2020}. Since the performance of ML models heavily relies on the quality and representativeness of training data, models trained on specific datasets may not generalise well to new or more diverse datasets, leading to reduced robustness and reliability in real-world applications. Variations in data collection protocols, equipment, and participant populations can impact the performance and generalisability of the models. 
Therefore, benchmark datasets should be controlled for various influencing factors\eg~age, sex, body height, body mass, speed differences, in order to allow for unbiased training and evaluation of ML models that are able to cope with inter- and intra-subject variability observed in biomechanical data.
Large-scale benchmark datasets would allow for a more reliable assessment of the ML model performance on unseen test data\eg~from another laboratory. With a larger dataset, the utilisation of more complex ML models, such as Deep Learning architectures holds the potential to yield deeper insights and improve the performance of ML models. 
Furthermore, large-scale benchmark datasets would allow the application of transfer learning approaches. This is a commonly employed technique, particularly, used in the field of computer vision to achieve high generalisation ability despite limited datasets. Transfer learning involves pre-training ML models on large existing datasets and fine-tuning them on a smaller task-relevant dataset. An alternative strategy for addressing the limited size of datasets involves artificially expanding them through the use of Generative Adversarial Networks. These artificial networks can assist in data augmentation by learning the data distribution and enabling the generation of new training data. In the context of spinal posture data, Generative Adversarial Networks have demonstrated potential by enhancing the accuracy of deep Autoencoders through the incorporation of augmented data ~\cite{dindorf2024enhancing}.

Another major limitation relates to the availability of annotations. Accurate and comprehensive annotations of biomechanical data, whether in the context of walking, running or sports movements, is a labour-intensive and subjective process. Annotating data requires human domain experts to identify and label specific events and features in the data. This process can introduce variability. Thus, it is important to create standardised annotation protocols that can be consistently applied to different datasets and research studies. In order to mitigate the requirement for large volumes of annotated data, several promising approaches including unsupervised and self-supervised ML methods could be utilised. These methods allow models to learn a general and transferable representation from large-scale non-annotated data that can be subsequently refined for specific downstream tasks.

The most promising strategy for generating large-scale and comprehensive datasets involves combining existing datasets from various laboratories and studies. Ensuring standardised procedures for data collection, storage, and annotation is crucial to enable this integration of datasets acquired from different sources. This standardisation would foster interoperability among the data. A first step in this direction in the context of walking data has been initiated with the GaitRec dataset and the publication of a standardised data preprocessing pipeline~\cite{horsak.2020}. Data sharing initiatives are essential to gather datasets from multiple laboratories and real-world scenarios and thereby promote future research.

\subsection{Further Limitations}
\label{subsec:furtherlimitations}

\textbf{Lack of Reproducibility.} The current state of practice indicates that the level of reproducibility in the context of ML in biomechanics is restricted. A significant limitation in this regard is the aforementioned challenge regarding the availability of data (Section~\ref{subsec:datascarcity}). Furthermore, ML models are frequently not shared, and the accessibility of code is often lacking. To enhance the reproducibility and transparency of research, the implementation of a badging system for publications in biomechanics similar to the ACM Artifact Review and Badging approach \cite{artifact_review.2020} could prove highly beneficial. Such an initiative could play a pivotal role in promoting open science practices and fostering a culture of transparency and collaboration within the biomechanics community, ultimately advancing the field's credibility and impact.

\textbf{Lack of Comparability.} Another significant challenge in the domain of ML in biomechanics is the limited comparability among studies. This lack of comparability results from the substantial variation in the evaluation strategies utilised in this community. The assessment of ML model performance in biomechanics can exhibit considerable diversity. This diversity encompasses various aspects, including the selection of evaluation strategies\eg~hold-out, k-fold cross-validation, and leave-one-out cross-validation, the use of different evaluation metrics, and the lack of benchmark datasets, which could be used to additionally validate a proposed ML approach. This heterogeneity makes it difficult to draw meaningful comparisons between studies in the field. 
In response to this challenge, the creation of standardised evaluation protocols and metrics specifically designed for biomechanical applications could be immensely beneficial.
To address this challenge, efforts to establish standardised evaluation protocols and metrics tailored to the specific requirements of biomechanical applications\eg~imbalanced data and limited dataset size are essential. Such standardisation would not only enhance the credibility of research findings but also facilitate the synthesis of knowledge and the development of more robust and reliable ML solutions.

\textbf{Neglecting the Biomechanical Context.} Gait representations learned solely in a data-driven approach have shown certain limitations in accurately capture the gait primitives that underlie fundamental locomotion patterns. Addressing these challenges necessitates a shift towards more comprehensive approaches, using not only data-driven learning but also physics-informed ML approaches.

Physics-informed ML methods integrate domain-specific knowledge, such as biomechanical principles, into the ML model. These methods have the potential to allow a more accurate modelling of the underlying physics of the biomechanical signals. As a result, the ML model could exhibit greater generalisability.

\section{Conclusion - The future of Machine Learning in Biomechanics}
\label{sec:conclusion}

The gradual integration of ML into biomechanics marks a significant transformation in the study of human movements. Reflecting on the current research landscape, it is evident that the domains of both gait and sports biomechanics can substantially benefit from the application of ML. While ML has made substantial progress in various key applications, its integration into biomechanics lags behind domains such as healthcare. This disparity is particularly evident in sports biomechanics, where considerably less publications used ML approaches compared to gait biomechanics. Nevertheless, a reciprocal learning dynamic might emerge between gait and sports biomechanics, both in terms of potential insights and methodological advances. However, it should be noted that these two domains are not always easily distinguishable from each other, and there are transitional areas between them. 

In certain applications, gait biomechanics is ahead in the development of ML-based approaches, outpacing its implementation in sports biomechanics. Sports biomechanists can obtain guidance on data preprocessing procedures and ML methodologies from gait biomechanics, as a larger number of ML pipelines and systematic comparisons are already available.
This includes findings such as the benefits of incorporating entire time series data and dimension reduction techniques such as Principal Component Analysis for input feature generation, as opposed to solely relaying on handcrafted input features. The findings also indicate that ML methods like Support Vector Machines and Random Forests have exhibited comparable, or sometimes even superior, performance to Deep Learning approaches and might be particularly suitable for small-scale datasets in biomechanics. Furthermore, there exists a need to cultivate substantially larger publicly accessible datasets in sports biomechanics compared to already available open source datasets in gait biomechanics. 

In comparison, the realm of sports biomechanics has also demonstrated promising advancements in utilising ML, thereby offering potential ML approaches that can be applied to the study of gait biomechanics. For instance, sports-related scenarios frequently involve real-world field conditions, in comparison to the controlled laboratory settings commonly seen in current applications of ML in gait research. Gait biomechanics can get guidance from ML applications in complex sports movements for applications that should work reliably in everyday clinical life.

However, ML in gait and sports biomechanics share similar inherent limitations. Addressing these challenges related to the accessibility of large benchmark datasets or to issues of variability, transparency, and trustworthiness, will require collaborative efforts. In order to enrich the community in a sustainable and future-oriented way, we need the willingness to share data and work together to advance new developments and achieve greater consensus regarding the application and evaluation of the ML methods. As we navigate this dynamic landscape, interdisciplinary collaboration will serve as the compass guiding us toward unlocking the full potential that ML offers in the multifaceted domains of gait and sports biomechanics.

\section*{Acknowledgements}
This work was partly funded by the Research Promotion Agency of Lower Austria (Gesellschaft für Forschungsförderung NÖ) within the Endowed Professorship for Applied Biomechanics and Rehabilitation Research (\#SP19-004) and the project DeepForce (\#GLF21-1-009), as well as by the Austrian Research Promotion Agency (FFG) within the COIN-program (\#898085). This research was also supported by the Central Innovation Programme for Small and Medium-Sized Enterprises (Zentrales Innovationsprogramm Mittelstand, ZIM) of the German Federal Ministry for Economic Affairs and Climate Action, under grant number 16KN113027.

\bibliographystyle{ieeetr}  
\bibliography{Refs}

\end{document}